\documentclass[10pt,journal,twocolumn]{IEEEtran}

\usepackage[nocompress]{cite}

\usepackage{times}
\usepackage{epsfig}
\usepackage{graphicx}
\usepackage{amsmath}
\usepackage{amssymb}
\usepackage{eucal}
\usepackage{cite}
\usepackage{cs}
\usepackage{mathsymb}
\usepackage{textcomp}
\usepackage{verbatim}
\usepackage{subfigure}
\usepackage{url}
\usepackage{multirow}
\usepackage[super]{nth}
\usepackage{colortbl}
\graphicspath{{./}{./Fig/}{./Figs/}{./Fig/eps/}{./Fig/pdf/}}

\usepackage{dblfloatfix}

\usepackage{wrapfig}
\usepackage{caption}
\usepackage[linesnumbered,algoruled,boxed,lined]{algorithm2e}

\def\be{{\boldsymbol e}}
\def\bx{{\boldsymbol x}}
\def\bz{{\boldsymbol z}}
\def\bS{{\boldsymbol S}}
\def\bX{{\boldsymbol X}}

\newcommand{\algorithmicrequire}{\textbf{Input: }}
\newcommand{\algorithmicensure}{\textbf{Output: }}

\let\savedalgorithm\algorithm
\let\savedendalgorithm\endalgorithm

\usepackage{algorithm}

\def\etal{\textit{et al.\@\xspace}}
\newcommand{\eg}{\textit{e.g.\@\xspace}}
\newcommand{\ie}{\textit{i.e.\@\xspace}}

\begin{document}

\title{Deep Linear Discriminant Analysis on Fisher Networks: A Hybrid Architecture for Person Re-identification}

\author{Lin Wu, Chunhua Shen, Anton van den Hengel %
\IEEEcompsocitemizethanks{\IEEEcompsocthanksitem Authors are with The University of
Adelaide, Australia; and Australian Research Council Centre of Excellence
for Robotic Vision;
Corresponding author: Chunhua Shen (e-mail: chunhua.shen@adelaide.edu.au).\protect\\
}

}

\IEEEtitleabstractindextext{%

\begin{abstract}

Person re-identification is to seek a correct match for a person of interest across views among a large number of imposters.
It typically involves two procedures of non-linear feature  extractions against dramatic appearance changes, and subsequent discriminative analysis in order to reduce intra-personal variations while enlarging inter-personal differences.
In this paper, we introduce a hybrid architecture which combines Fisher vectors and deep neural networks to learn non-linear
representations of person images to a space where data can be linearly separable.
We reinforce a Linear Discriminant Analysis (LDA) on top of the deep neural network such that linearly separable latent representations can be learnt in an end-to-end fashion.

By optimizing an objective function modified from LDA, the network is enforced to produce feature distributions which have a low variance within the same class and high variance between classes. The objective is essentially derived from the general LDA eigenvalue problem and allows to train the network with stochastic gradient descent and back-propagate LDA gradients to compute the gradients involved in Fisher vector encoding.
For evaluation we test our approach on four benchmark data sets in person re-identification
(VIPeR \cite{Gray2007Evaluating}, CUHK03 \cite{FPNN}, CUHK01 \cite{CUHK01}, and Market1501 \cite{Market1501}).
Extensive experiments on these benchmarks show that our model can achieve state-of-the-art results.

\end{abstract}

\begin{IEEEkeywords}
  Linear discriminant analysis, deep Fisher networks, person re-identification.
\end{IEEEkeywords}

}

\maketitle

\IEEEdisplaynontitleabstractindextext

\IEEEpeerreviewmaketitle

\section{Introduction}\label{sec:intro}

\begin{figure}[t]
\centering
\includegraphics[height=5cm]{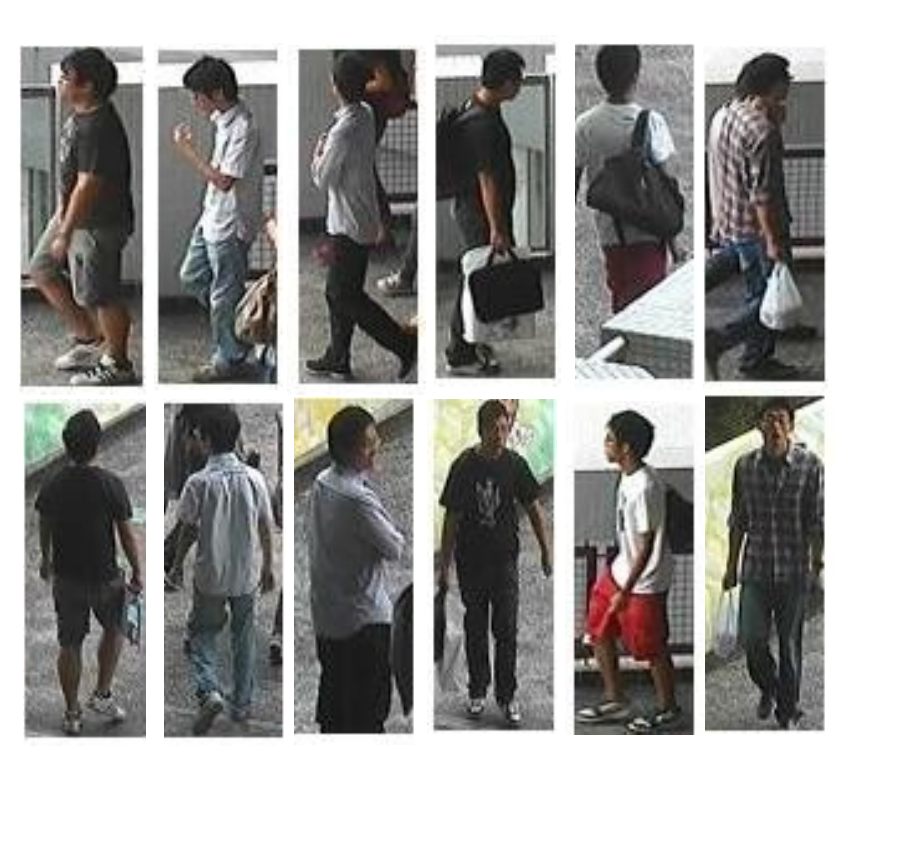}
\caption{Samples of pedestrian images observed in different camera views in CUHK03 data set \cite{DeepReID}. Images in columns display the same identity.}
\label{fig:examples}
\end{figure}

\IEEEPARstart{T}{he} problem of person re-identification (re-id) refers to matching pedestrian images observed from disjoint camera views
(some samples are shown in Fig.~\ref{fig:examples}).
Many approaches to this problem have been proposed mainly to develop discriminative feature representations that are robust against poses/illumination/viewpoint/background changes
\cite{Cheng2011Custom,Farenzena2010Person,Gheissari2006Person,MidLevelFilter,Gray2008Viewpoint,eSDC},
or to learn distance metrics for matching people across views \cite{Davis2007Information,LADF,Multi-taskDistance,Kostinger2012Large,Wu2011Optimizing,Weinberger2006Distance,PCCA,Zheng2013PAMI,Xiong2014Person,LOMOMetric,RankSVM,paul2015ensemble}, or both jointly \cite{FPNN,PersonNet,JointRe-id,DeepReID}.

To extract non-linear features against complex and large visual appearance variations in person re-id, some deep models based on Convolution Neural Networks (CNNs)
\cite{JointRe-id,PersonNet,FPNN,DeepReID,DeepRanking} have been developed to compute robust data-dependent feature representations. Despite the witnessed improvment achieved by deep models, the learning of parameters requires a large amount of training data in the form of matching pairs whereas person re-id is facing the small sample size problem \cite{SSSProblem,NullSpace-Reid}. Specifically, only hundreds of training samples are available due to the difficulties of collecting matched training images. Moreover, the loss function used in these deep models is at the
single image level. A popular choice is the cross-entropy loss which is used to maximize the probability of each pedestrian image. However, the gradients computed from the class-membership probabilities may help enlarging the inter-personal differences while unable to reduce the intra-personal variations. To these ends, we are motivated to develop a architecture that can be comparable to deep convolution but more suitable for person re-identification with moderate data samples. At the same time, the proposed network is expected to produce feature representations which could be easily separable by a linear model in its latent space such that visual variation within the same identity is minimized while visual differences between identities are maximized.

Fisher kernels/vectors and CNNs can be related in a unified view to which there are much conceptual similarities between both architectures instead of
structural differences \cite{DeepFisherKernel}. Indeed, the encoding/aggregation step of Fisher vectors which typically involves gradient filter,
pooling, and normalization can be interpreted as a series of layers that alternate linear and non-linear operations, and hence can be paralleled with the convolutional layers of CNNs.
Some recent studies \cite{FVsMeetNNs,DeepFisherNetworks,DeepFisherKernel} have empirically demonstrated that Fisher vectors are competitive with representations
learned by convolutional layers of CNNs, and the accuracy of Fisher vector comes very close to that of the AlexNet \cite{AlexNet} in image classification.
For instance, Sydorov \etal \cite{DeepFisherKernel} transfer the idea of end-to-end training and discriminatively learned feature representations in neural networks to
support vector machines (SVMs) with Fisher kernels. As a result, Fisher kernel SVMs can be trained in a deep way by which classifier parameters
and Gaussian mixture model (GMM) parameters are learned jointly from training data.
The resulting Fisher kernels can provide state-of-the-art results in image classification \cite{DeepFisherKernel}. While complex CNN models require a vast number of training data to find good representations and task-dependent parameters, learning  GMM parameters in Fisher kernels usually only demands a moderate amount of training data due to the nature of unsupervised learning.
Thus, we are interested in developing a deep network based on Fisher vectors, which can learn effective features to match persons across views.

Person re-identification can be reformulated into a process of  discriminant analysis on a lower dimensional space into which training data of matching image pairs across
camera views are projected. For instance, Pedagadi \etal \cite{Pedagadi2013Local} presented a method with Fisher Discrimant Analysis (FDA) where discriminative subspace is formed from learned discriminative projecting directions, on which within (between)-class distance is minimized (maximized). They exploited graph Laplacian to preserve local data structure, known as LFDA.
Also, Zhang \etal \cite{NullSpace-Reid} proposed to learn a discriminative null space for person re-id where the FDA criterion can be satisfied to the extreme
by collapsing images of the same person into a single point.

\begin{figure*}[t]
\centering
\includegraphics[height=5cm]{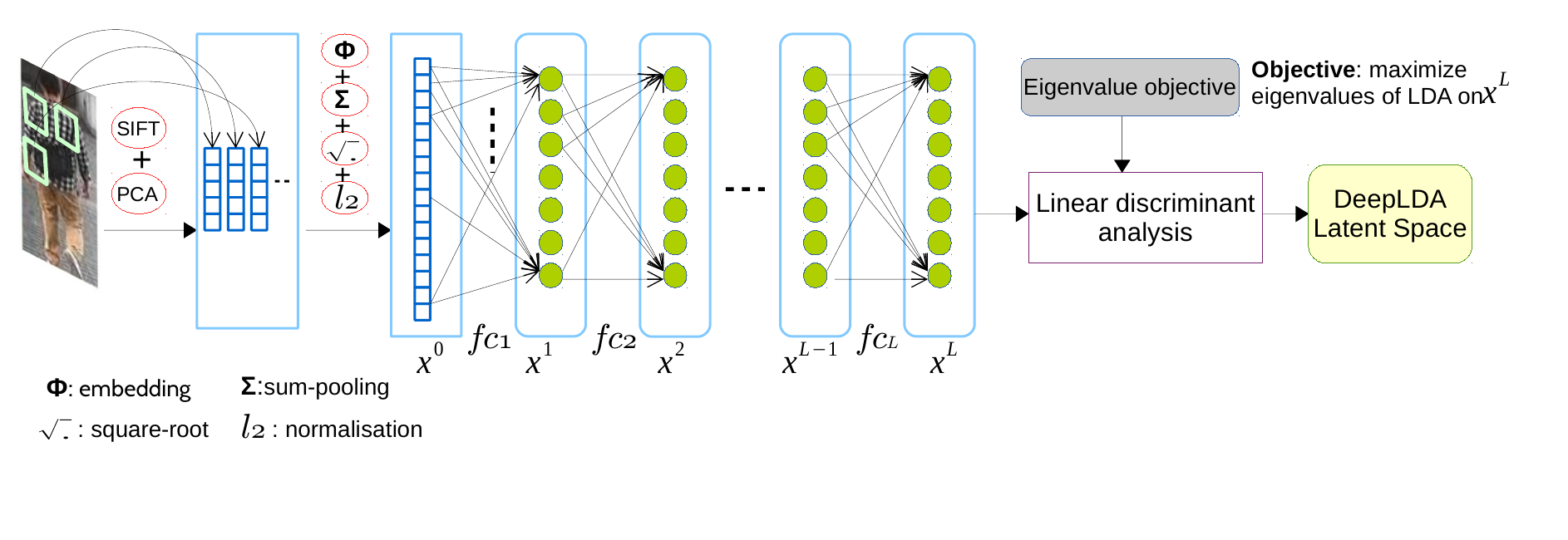}
\caption{Overview of our architecture. Patches from a person image are extracted and described by PCA-projected SIFT descriptors. These local descriptors are further embeded using Fisher vector encoding and aggregated to form its image-level representation, followed by square-rooting and $\ell_2$-normalization ($x^0$). Supervised layers $fc_1,\dots,fc_L$ that involve a linear projection and an ReLU can transform resulting Fisher vectors into a latent space into which an LDA objective function is directly imposed to preserve class separability.}
\label{fig:framework}
\end{figure*}

In this paper, we present a hybrid architecture for person re-identification, which is comprised of Fisher vectors and multiple supervised layers. The network is trained with an Linear Discriminant Analysis (LDA) criterion because LDA approximates inter- and intra-class variations by using two scatter matrices and finds the projections to maximize the ratio between them. As a result, the computed deeply non-linear features become linearly separable in the resulting latent space. The overview of architecture is shown in Fig.\ref{fig:framework}.

The motivation of our architecture is three-fold. First, we expect to seek alternative to CNNs (\eg, Fisher vectors)
while achieving comparable performance such that the training computational cost is reduced substantially and the over-fitting
issue may be alleviated.
Second, we would like to see whether the discriminative capability of Fisher vectors can be improved in preserving class separability
if LDA gradients are back propagated to compute GMM gradients. Third, in the deep latent linearly separable feature space
induced by non-linear LDA, we want to exploit the beneficial properties of LDA such as low intra-class variance,
high inter-class variance, and optimal decision boundaries, to gain  plausible improvements for the task of person re-identification.

\subsection{Our method}
The proposed hybrid architecture is a feed-forward neural network. The network feed-forwardly encodes the SIFT features through
a parametric generative model, \ie\ Gaussian Mixture Model (GMM), a set of fully connected supervised layers, followed by a modified LDA objective function.
Then, the parameters are optimized by back-propagating the error of an LDA-based objective through the entire network such that GMM parameters are updated to have task-dependent values.  Thus, the established non-linearities of Fisher kernels provide a natural way to extract good discriminative
features from raw measurements.
For LDA, since discriminative features generated from non-linear transformations of mixture Gaussian distribution, it is capable of finding linear combinations of the input features which allows for optimal linear decision boundaries.
To validate the proposed model, we apply it to four popular
benchmark datasets: VIPeR \cite{Gray2007Evaluating}, CUHK03 \cite{DeepReID}, CUHK01 \cite{CUHK01}, and Market-1501 \cite{Market1501},
which are moderately sized data sets.
From this perspective, deep Fisher kernel learning can keep the number of parameters reasonable so as to enable the training on medium sized data sets while achieving competitive results.

Our major contributions can be summarized as follows.

\begin{itemize}
  \item A hybrid deep network comprised of Fisher vectors, stacked of fully connected layers, and an LDA objective function is presented.
    The modified LDA objective allows to train the network in an end-to-end manner where LDA derived gradients are back-propagated to update GMM parameters.
  \item The beneficial properties of LDA are embedded into deep network to directly produce linearly separable feature representations in the resulting latent space.
\item Extensive experiments on benchmark data sets are conducted to show our model is able to achieve state-of-the-art performance.
\end{itemize}

The reminding part of the paper is structured as follows. In  Section \ref{sec:related}  we review related works.
In Section \ref{sec:approach}, we introduce linear discriminant analysis on deep Fisher kernels, a hybrid system that learns linearly separable latent representations in an end-to-end fashion. Then in Section \ref{sec:relationship} we explain how the proposed architecture relates to deep Fisher kernel learning and CNNs. In Section \ref{sec:exp}, we experimentally evaluate the proposed model to benchmarks of person re-id. Section \ref{sec:con} concludes this paper. %

\section{Related work}\label{sec:related}

\subsection{Person re-identification}

Recent studies on person re-id primarily focus on developing new feature descriptors or learning optimal distance metrics across views or both jointly.

Many person re-id approaches aim to seek robust and discriminative features that can describe the appearance of the same individual across different camera views under various changes and conditions \cite{Cheng2011Custom,Farenzena2010Person,Gheissari2006Person,MidLevelFilter,Gray2008Viewpoint,eSDC}.
For instance, Gray and Tao introduced an ensemble of local features which combines three color channels with 19 texture channels \cite{Gray2008Viewpoint}.
Farenzena  \etal\
\cite{Farenzena2010Person} proposed the Symmetry-Driven Accumulation of Local Features (SDALF) that exploited the symmetry property of a person where positions of head, torso, and legs were used to handle view transitions. Zhao \etal\
combined color SIFT features with unsupervised salience learning to improve its discrminative power in person re-id (known as eSDC) \cite{eSDC},
and further integrated both salience matching and patch matching into an unified RankSVM framework (SalMatch \cite{Zhao2013SalMatch}).
They also proposed mid-level filters for person re-id by exploring the partial area under the ROC curve (pAUC) score \cite{MidLevelFilter}.
However, descriptors of visual appearance are highly susceptible to cross-view variations due to the inherent visual ambiguities and disparities caused by different view orientations, occlusions, illumination, background clutter, etc.
Hence, it is difficult to strike a balance between discriminative power and robustness. In addition, some of these methods heavily rely on foreground segmentations, for instance, the method of SDALF \cite{Farenzena2010Person} requires high-quality silhouette masks for symmetry-based partitions.

Metric learning approaches to person re-id work by extracting features for each image first, and then learning a metric with which the training data have strong inter-class differences and intra-class similarities. In particular, Prosser \etal\ \cite{RankSVM} developed an ensemble RankSVM to learn a subspace where the potential true match is given the highest ranking. Koestinger \etal\ \cite{Kostinger2012Large} proposed the large-scale metric learning from equivalence constraint (KISSME) which considers a log likelihood ratio test of two Gaussian distributions. In \cite{LADF}, a Locally-Adaptive Decision Function is proposed to jointly learn distance metric and a locally adaptive thresholding rule. An alternative approach is to use a logistic function to approximate the hinge loss so that the global optimum still can be achieved by iteratively gradient search along the projection matrix as in PCCA \cite{PCCA}, RDC \cite{Zheng2013PAMI}, and Cross-view Quadratic Discriminant Analysis (XQDA) \cite{LOMOMetric}. Pedagadi \etal \cite{Pedagadi2013Local} introduced the Local Fisher Discriminant Analysis (LFDA) to learn a discriminant subspace after reducing the dimensionality of the extracted features. A further kernelized extension (kLFDA) is presented in \cite{Xiong2014Person}. However, these metric learning methods share a main drawback that they need to work with a reduced dimensionality, typically achieved by PCA. This is because the sample size of person re-id data sets is much smaller than feature dimensionality, and thus metric learning methods require dimensionality reduction or regularization to prevent matrix singularity in within-class scatter matrix \cite{LOMOMetric,Pedagadi2013Local,Xiong2014Person}. Zhang \etal\ \cite{NullSpace-Reid} addressed the small sample size problem by matching people in a discriminative null space in which images of the same person are collapsed into a single point. Nonetheless, their performance is largely limited by the representation power of hand-crafted features, \ie, color, HOG, and LBP.

More recently, deep models based on CNNs that extract features in hierarchy have shown great potential in person re-id. Some recent representatives include FPNN \cite{FPNN}, PersonNet \cite{PersonNet}, DeepRanking \cite{DeepRanking}, and Joint-Reid \cite{JointRe-id}. FPNN is to jointly learn the representation of a pair of person image and their corresponding metric. An improved architecture was presented in \cite{JointRe-id} where they introduced a layer that computes cross-input neighborhood difference featuers immediately after two layers of convolution and max pooling. Wu \etal\ \cite{PersonNet} improved the architecture of Joint-Reid \cite{JointRe-id} by using more deep layers and smaller convolutional filter size. However, these deep models learn a network with a binary classification, which tends to predict most input pairs as negatives due to the great imbalance of training data. To avoid data imbalance, Chen \etal\ \cite{DeepRanking} proposed a unified deep learning-to-rank framework (DeepRanking) to jointly learn representation and similarities of image pairs.

\subsection{Hybrid systems}
Simonyan \etal\ \cite{DeepFisherNetworks} proposed a deep Fisher network by stacking Fisher vector image encoding into multiple layers. This hybrid architecture can significantly improve on standard Fisher vectors, and obtain competitive results with deep convolutional networks at a smaller computational cost.
A deep Fisher kernel learning method is presented by Sydorov \etal\ \cite{DeepFisherKernel} where they improve on Fisher vectors by jointly learning the SVM classifier and the GMM visual vocabulary. The idea is to derive SVM gradients which can be back-propagated to compute the GMM gradients. Finally, Perronnin \cite{FVsMeetNNs} introduce a hybrid system that combines Fisher vectors with neural networks to produce representations for image classifiers. However, their network is not trained in an end-to-end manner.

\section{Our architecture}\label{sec:approach}

\begin{figure}[hbt]
\centering
\includegraphics[height=5cm]{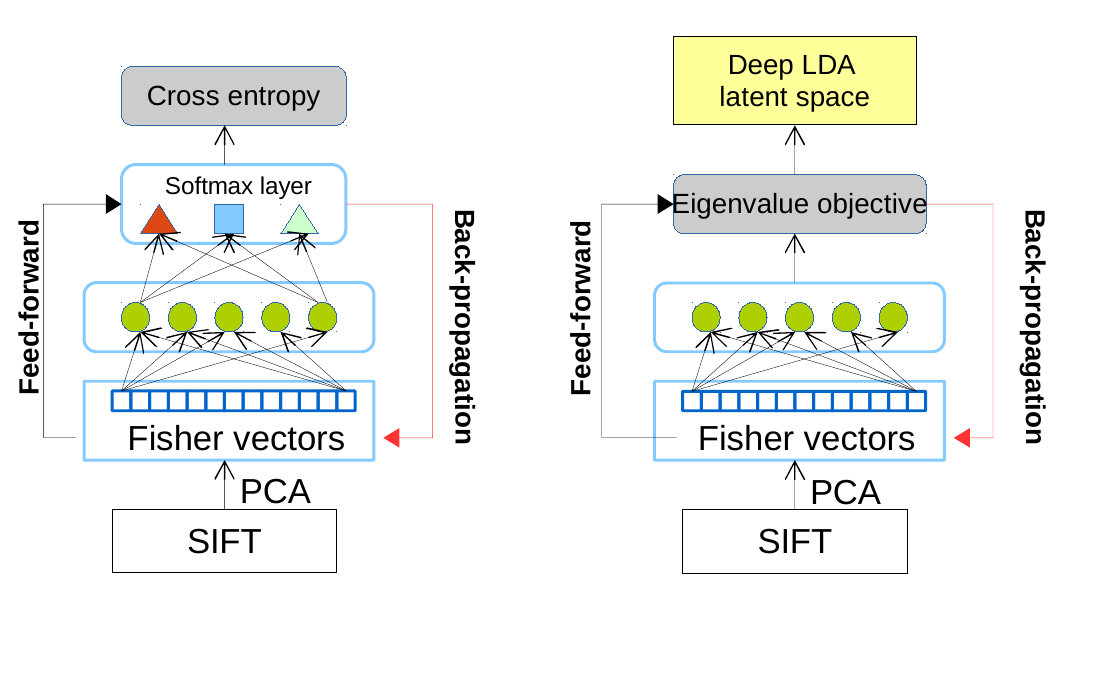}
\caption{ Sketch of a general DNN and our deep hybrid LDA network. In a general DNN, the outputs of network are normalized by a softmax layer to obtain probabilities and the objective with cross-entropy loss is defined on each training sample.  In our network, an LDA objective is imposed on the hidden representations, and the optimization target is to directly maximize the separation between classes.}
\label{fig:LDA_class}
\end{figure}

\subsection{Fisher vector encoding}

The Fisher vector encoding $\Phi(\cdot)$ of a set of features is based on  fitting a parametric generative model such as the Gaussian Mixture Model (GMM) to the features, and then encoding the derivatives of the log-likelihood of the model with respect to its parameters \cite{GenerativeNIPS1998}. It has been shown to be a state-of-the-art local patch encoding technique \cite{FisherVectorIJCV,DevilBMVC2011}. Our base representation of an image is a set of local descriptors, \eg, densely computed SIFT features. The descriptors are first PCA-projected to reduce their dimensionality and decorrelate their coefficients to be amenable to the FV description based on diagonal-covariance GMM.
Given a GMM with $K$ Gaussians, parameterized by $\{\pi_k, \mu_k, \sigma_k, k=1\ldots, K\}$, the Fisher vector encoding leads to the representation which captures the average first and second order differences between the features and each of the GMM centres \cite{ImproveFisherKernel,DeepFisherNetworks}. For a descriptor $x\in \mathbb{R}^D$, we define a vector $\Phi(x)=\left[\varphi_1(x),\ldots, \varphi_K(x), \psi_1(x),\ldots,\psi_K(x)\right]\in \mathbb{R}^{2KD}$. The subvectors are defined as
\begin{equation}\label{eq:FV}
\begin{split}
&\varphi_k (x)=\frac{1}{\sqrt{\pi_k}} \gamma_k (x) \left(\frac{x-\mu_k}{\sigma_k}\right)\in \mathbb{R}^D,\\
& \psi_k(x)=\frac{1}{\sqrt{2\pi_k}}\gamma_k(x) \left(\frac{(x-\mu_k)^2}{\sigma_k^2}-1\right) \in \mathbb{R}^D,
\end{split}
\end{equation}
where $\{\pi_k, \mu_k, \sigma_k\}_k$ are the mixture weights, means, and diagonal covariance of the GMM, which are computed on the training set; $\gamma_k(x)$ is the soft assignment weight of the feature $x$ to the $k$-th Gaussian. In this paper, we use a GMM with $K=256$ Gaussians.

To represent an image $X=\{x_1,\ldots,x_M\}$, one averages/sum-pool the vector representations of all descriptors, that is, $\Phi(X)=\frac{1}{M}\sum_{i=1}^M \Phi(x_i)$. $\Phi(X)$ refers to Fisher vector of the image since for any two image $X$ and $X'$, the inner product $\Phi(X)^T\Phi(X')$ approximates to the Fisher kernel, $k(X,X')$, which is induced by the GMM. The Fisher vector is further processed by performing signed square root and $\ell_2$ normalization on each dimension $d=1,\ldots, 2KD$:
\begin{equation}\label{eq:fisher_norm}
  \bar{\Phi}_d(X)=(\sign \cdot \Phi_d(X)) \sqrt{|\Phi_d(X)|}/ \sqrt{||\Phi(X)||_{\ell_1}}.
\end{equation}
Here $ \sign $ is the sign of the original vector.
For simplicity, we refer to the resulting vectors from Eq.~\eqref{eq:fisher_norm} as Fisher vectors and to their inner product as Fisher kernels.

\subsection{Supervised layers}

In our network, the PCA-reduced Fisher vector outputs are fed as inputs to a set of $L$ fully connected supervised layers $fc_1,\dots,fc_L$. Each layer involves a linear projection followed by a non-linear activation.
Let $x^{l-1}$ be the output of layer $fc_{l-1}$, and $\sigma$ denotes the non-linearity, then we have the output $x^l=\sigma(F^l (x^{l-1}))$ where $F^l(x) = W^l x + b^l$, and $W^l$ and $b^l$ are parameters to be learned. We use a rectified Linear Unit (ReLU) non-linearity function $\sigma(x)=\max(0,x)$, which showed improved convergence compared with sigmoid non-linearities \cite{AlexNet}.
As for the output of the last layer $fc_L$, existing deep networks for person re-id commonly use a softmax layer which can derive a probabilistic-like output vector against each class. Instead of maximizing the likelihood of the target label for each sample, we use deep network to learn non-linear transforms of training samples to a space in which these samples can be linearly separable. In fact, deep networks are propertied to have the ability to concisely represent a hierarchy of features for modeling real-world data distributions, which could be very useful in a setting where the output space is more complex than a single label.

\subsection{Linear discriminant analysis on deep Fisher networks}

Given a set of $N$ samples $\bX=\{\bx_1^L,\ldots,\bx_N^L\}\in \mathbb{R}^{N\times d}$ belonging to different classes $\omega_i$, $1 \leq i\leq C$, where $\bx_i^L$ denotes output features of a sample from the supervised layer $fc_L$, linear discriminant analysis (LDA) \cite{LDA} seeks a linear transform $\textbf{A}: \mathbb{R}^d \rightarrow \mathbb{R}^l$, improving the discrimination between features $\bz_i=\bx_i^L \textbf{A}^T$ lying in a lower $l$-dimensional subspace $L \in \mathbb{R}^l$ where $l=C-1$. Please note that the input representation $\bX$ can also be referred as hidden representation. Assuming discriminative features $\bz_i$ are identifically and independently drawn from Gaussian class-conditional distributions, the LDA objective to find the projection matrix \textbf{A} is formulated as:
\begin{equation}\label{eq:ratio}
\arg\max_\textbf{A} \frac{|\textbf{A} \bS_b \textbf{A}^T|}{|\textbf{A} \bS_w \textbf{A}^T|},
\end{equation}
where $\bS_b$ is the between scatter matrix, defined as $\bS_b=\bS_t-\textbf{S}_w$. $\bS_t$ and $\bS_w$ denote total scatter matrix and within scatter matrix, which can be defined as
\begin{equation}
\begin{aligned}
& \bS_w =\frac{1}{C}\sum_c \bS_c,  \bS_c = \frac{1}{N_c -1}\bar{\bX}_c ^T \bar{\bX}_c; \\
& \bS_t = \frac{1}{N-1} \bar{\bX}^T \bar{\bX};
\end{aligned}
\end{equation}
where $\bar{\bX}_c = \bX_c - \textbf{m}_c$ are the mean-centered observations of class $\omega_c$ with per-class mean vector $\textbf{m}_c$. $\bar{\bX}$ is defined analogously for the entire population $\bX$.

In \cite{GerDA}, the authors have shown that nonlinear transformations from deep neural networks can transfrom arbitrary data distributions to Gaussian distributions such that the optimal discriminant function is linear. Specifically, they designed a deep neural network consituted of unsupervised pre-optimization based on stacked Restricted Boltzman machines (RBMs) \cite{RBMs} and subsequent supervised fine-tuning. Similarly, the Fisher vector encoding part of our network is a generative Gaussian mixture mode, which can not only handle real-value inputs to the network but also model them continuously and amenable to linear classifiers. In some extent, one can view the first layers of our network as an unsupervised stage but these layers need to be retrained to learn data-dependent features.

 Assume discriminant features $\bz=\textbf{A}\bx$, which are identically and independently drawn from Gaussian class-conditional distribution, a maximum likelihood estimation of \textbf{A} can be equivalently computed by a maximation of a discriminant criterion: $Q_z: \mathcal{F} \rightarrow \mathbb{R}, \mathcal{F}=\{\textbf{A}: \textbf{A}\in \mathbb{R}^{l\times d}\}$ with $Q_z (\textbf{A})=\max (trace\{ \bS_t^{-1} \bS_b\} )$. We use the following theorem from the statistical properties of LDA features to obtain a guarantee on the maximization of the discriminant criterion $Q_z$ with the learnt hidden representation $\bx$.

\paragraph{Corollary} On the top of the Fisher vector encoding, the supervised layers learn input-output associations leading to an indirectly maximized $Q_z$. In \cite{OsmanPAMI1994}, it has shown that multi-layer artificial neural networks with linear output layer
\begin{equation}
\bz_n = \textbf{W}\bx_n + \textbf{b}
\end{equation}
maximize asymptotically a specific discriminant criterion evaluated in the $l$-dimensional space spanned by the last hidden outputs $\bz\in \mathbb{R}^l$ if the mean squared error (MSE) is minimized between $\bz_n$ and associated targets $\textbf{t}_n$. In particular, for a finite sample $\{\bx_n\}_{n=1}^N$, and MSE miminization regarding the target coding is
\[ t_n^i = \left\{\begin{array}{cl}
\sqrt{N/N_i}, & \omega(\bx_n) = \omega_i \\
0,& \mbox{otherwise}\end{array}\right.\]
where $N_i$ is the number of examples of class $\omega_i$, and $t_n^i$ is the $i$-th component of a target vector $\textbf{t}_n \in \mathbb{R}^C$, approximates the maximum of the discriminant criterion $Q_h$.

Maximizing the objective function in Eq.~\eqref{eq:ratio} is essentially to maximize the ratio of the between and within-class scatter, also known as separation. The linear combinations that maximize Eq.\eqref{eq:ratio} leads to the low variance in projected observations of the same class, whereas high variance in those of different classes in the resulting low-dimensinal space $L$. To find the optimum solution for Eq.\eqref{eq:ratio}, one has to solve the general eigenvalue problem of $\bS_b \be= \textbf{v} \bS_w \be$, where $\be$ and \textbf{v} are eigenvectors and corresponding eigenvalues. The projection matrix \textbf{A} is a set of eigenvectors $\be$.
In the following section, we will formulate LDA as an objective function for our hybrid architecture, and present back-propagation in supervised layers and Fisher vectors.

\subsection{Optimization}\label{ssec:optimization}

Existing deep neural networks in person re-identification \cite{JointRe-id,FPNN,PersonNet} are optimized using sample-wise optimization with cross-entropy loss on the predicted class probabilities (see Section below). In constrast, we put an LDA-layer on top of the neural networks by which we aim to produce features with low intra-class and high inter-class variability rather than penalizing the misclassification of individual samples. In this way, optimization with LDA objective operates on the properties of the parameter distributions of the hidden representation generated by the hybrid net. In the following, we first briefly revisit deep neural nettworks with cross-entropy loss, then we present the optimization problem of LDA objective on the deep hybrid architecture.

\subsubsection{Deep neural networks with cross-entropy loss}

It is very common to see many deep models (with no exception on person re-identification) are built on top of a deep neural network (DNN) with training used for classification problem \cite{DeepFisherKernel,FVsMeetNNs,AlexNet,VGG}. Let $\mathcal{X}$ denote a set of $N$ training samples $X_1,\dots,X_N$ with corresponding class labels $y_1,\dots,y_n \in \{1,\dots,C\}$. A neural network with $P$ hidden layers can be represented as a non-linear function $f(\Theta)$ with model parameters $\Theta = \{\Theta_1,\dots,\Theta_P\}$. Assume the network output $\textbf{p}_i = [p_{i,1},\dots,p_{i,C}]=f(X_i,\Theta)$ is normalized by the softmax function to obtain the class probabilities, then the network is optimized by using Stochastic Gradient Descent (SGD) to seek optimal parameter $\Theta$ with respect to some loss function $\mathcal{L}_i (\Theta) = \mathcal{L}(f(X_i,\Theta), y_i)$:
\begin{equation}
\Theta = \arg \min_{\Theta} \frac{1}{N} \sum_{i=1}^N \mathcal{L}_i (\Theta).
\end{equation}

For a multi-class classification setting, cross-entropy loss is a commonly used optimization target which is defined as
\begin{equation}\label{eq:cross-entropy}
\mathcal{L}_i = - \sum_{j=1}^C y_{i,j} \log (p_{i,j}),
\end{equation}
where $y_{i,j}$ is 1 if sample $X_i$ belongs to class $y_i$ \ie ($j=y_i$), and 0 otherwise. In essence, the cross-entropy loss tries to maximize the likelihood of the target class $y_i$ for each of the individual sample $X_i$ with the model parameter $\Theta$. Fig.\ref{fig:LDA_class} shows the snapshot of a general DNN with such loss function. It has been emphasized in \cite{DeepLDA} that objectives such as cross-entropy loss do not impose direct constraints such as linear separability on the latent space representation. This movitates the LDA based loss function such that hidden representations are optimized under the constraint of maximizing discriminative variance.

\subsubsection{Optimization objective with linear discriminant analysis}

A problem in solving the general eigenvalue problem of $\bS_b \be= \textbf{v} \bS_w \be$ is the estimation of $\bS_w$ which overemphasises high eigenvalues whereas small ones are underestimated too much. To alleviate this effect, a regularization term of identity matrix is added into the within scatter matrix: $\bS_w + \lambda \textbf{I}$ \cite{Friedman1989}. Adding this identify matrix can stabilize small eigenvalues. Thus, the resulting eigenvalue problem can be rewritten as
\begin{equation}\label{eq:eigen_prob}
\bS_b \be_i = v_i (\bS_w + \lambda \textbf{I}) \be_i,
\end{equation}
where $\be=[\be_1,\dots,\be_{C-1}]$ are the resulting eigenvectors, and $\textbf{v} =[v_1,\dots,v_{C-1}]$ the corresponding eigenvalues. In particular, each eigenvalue $v_i$ quantifies the degree of separation in the direction of eigenvector $\be_i$.  In \cite{DeepLDA}, Dorfer \etal adapted LDA as an objective into a deep neural network by maximizing the individual eigenvalues. The rationale is that the maximization of individual eigenvalues which reflect the separation of respective eigenvector directions leads to the maximization of the discriminative capability of the neural net. Therefore, the objective to be optimized is formulated as:
\begin{equation}\label{eq:obj}
\arg \max_{\Theta, G} \frac{1}{C-1} \sum_{i=1}^{C-1} v_i.
\end{equation}
However, directly solving the problem in Eq.~\eqref{eq:obj} can yield trival solutions such as only maximizing the largest eigenvalues since this will produce the highest reward. In perspective of classification, it is well known that the discriminant function overemphasizes the large distance between already separated classes, and thus causes a large overlapping between neighboring classes \cite{GerDA}. To combat this matter, a simple but effective solution is to focus on the optimization on the smallest (up to $m$) of the $C-1$ eigenvalues \cite{DeepLDA}. This can be implemented by concentrating only $m$ eigenvalues that do not exceed a predefined threshold for discriminative variance maximation:
\begin{equation}\label{eq:modified_obj}
\begin{split}
&\mathcal{L}(\Theta, G) = \arg \max_{\Theta, G} \frac{1}{m} \sum_{i=1}^m v_i, \\
& \mbox{with} ~~\{v_1,\dots,v_m\}=\{v_i | v_i < \min \{v_1,\dots,v_{C-1} \} + \epsilon \}.
\end{split}
\end{equation}
The intuition of formulation \eqref{eq:modified_obj} is the learnt parameters $\Theta$ and $G$ are encouraged to embed the most discriminative capablility into each of the $C-1$ feature dimensions. This objective function allows to train the proposed hybrid architecture in an end-to-end fashion. In what follows, we present the derivatives of the optimization in \eqref{eq:modified_obj} with respect to the hidden representation of the last layer, which enables back-propagation to update $G$ via chain rule.

\paragraph{Gradients of LDA loss}

In this section, we provide the partial derivatives of the optimization function \eqref{eq:modified_obj} with respect to the output hidden representation $\bX\in \mathbb{R}^{N\times d}$ where $N$ is the number of training samples in a mini-batch and $d$ is the feature dimension output from the last layer of the network. We start from the general LDA eigenvalue problem of $\bS_b \be_i = v_i \bS_w \be_i$, and the derivative of eigenvalue $v_i$ w.r.t. $\bX$ is defined as \cite{Leeuw2007}:
\begin{equation}
\frac{\partial v_i}{\partial \bX} = \be_i^T \left( \frac{\partial \bS_b}{\partial \bX} - v_i \frac{\partial \bS_w}{\partial \bX} \right) \be_i.
\end{equation}
Recalling the definitions of $\bS_t$ and following \cite{GerDA,DeepCCA,DeepLDA},  we can first obtain the partial derivative of the total scatter matrix $\bS_t$ on $\bX$ as in Equation \eqref{equ_12}.
\begin{figure*}[h!]
\normalsize
\setcounter{equation}{11}
\begin{equation}
  \label{equ_12}
\frac{\partial \bS_t[a,b]}{\partial \bX[i,j]}=\left\{\begin{array}{cl}
 \frac{2}{N-1} (\bX[i,j] - \frac{1}{N} \sum_n \bX[n,j]), & \mbox{if} ~~a=j, b=j\\
 \frac{1}{N-1} (\bX[i,b] - \frac{1}{N} \sum_n \bX[n,b]), & \mbox{if} ~~a=j, b\neq j\\
 \frac{1}{N-1} (\bX[i,a] - \frac{1}{N} \sum_n \bX[n,a]), & \mbox{if} ~~a\neq j, b= j\\
0,& \mbox{if} ~~ a\neq j, b\neq j
\end{array}\right.
\end{equation}
\setcounter{equation}{12}
\hrulefill
\end{figure*}
Here $\bS_t[a,b]$ indicates the element in row $a$ and column $b$ in matrix $\bS_t$, and likewise for $\bX[i,j]$. Then, we can obtain the partial derivatives of $\bS_w$ and $\bS_b$ w.r.t. $\bX$:
\begin{equation}
\begin{split}
&\frac{\partial \bS_w[a,b]}{\partial \bX[i,j]} = \frac{1}{C} \sum_c \frac{\partial \bS_c[a,b]}{\partial \bX[i,j]}; \\
&\frac{\partial \bS_b[a,b]}{\partial \bX[i,j]} = \frac{1}{C} \sum_c \frac{\partial \bS_t[a,b]}{\partial \bX[i,j]} - \frac{\partial \bS_w[a,b]}{\partial \bX[i,j]}.
\end{split}
\end{equation}

Finally, the partial derivative of the loss function formulated in \eqref{eq:modified_obj} w.r.t. hidden state $\bX$ is defined as:
\begin{equation}
\frac{\partial}{\partial \bX}\frac{1}{m}\sum_{i=1}^m v_i =\frac{1}{m}\sum_{i=1}^m \frac{\partial v_i}{\partial \bX}=\frac{1}{m} \sum_{i=1}^m \be_i^T \left( \frac{\partial \bS_b}{\partial \bX} - \frac{\partial \bS_w}{\partial \bX}\right) \be_i.
\end{equation}

\paragraph{Backpropagation in Fisher vectors}

In this section, we introduce the procedure for the deep learning of LDA with Fisher vector.  Algorithm \ref{alg:algorithm1} shows the steps of updating parameters in the hybrid architecture. In fact, the hybrid network consisting of Fisher vectors and supervised layers updates its parameters by firstly initializing GMMs (line 5) and then computing $\Theta$ (lines 6-7). Thereafter, GMM parameters $G$ can be updated by gradient decent (line 8-12). Specifically, Line 8 computes the gradient of the loss function w.r.t.\ the GMM parameters $\pi$, $\mu$ and $\sigma$. Their influence on the loss is indirect through the computed Fisher vector, as in the case of deep networks, and one has to make use of the chain rule. Analytic expression of the gradients are provided in Appendix.

Evaluating the gradients numerically is computationally expensive due to the non-trivial couplings between parameters. A straight-forward implementation gives rise to complexity of $O(D^2 P)$ where $D$ is the dimension of the Fisher vectors, and $P$ is the combined number of SIFT descriptors in all training images. Therefore, updating the GMM parameters could be orders of magnitude slower than computing Fisher vectors, which has computational cost of $O(DP)$.
Fortunately, \cite{DeepFisherKernel} provides some schemes to compute efficient approximate gradients as alternatives.

It is apparent that the optimization w.r.t. $G$ is non-convex, and Algorithm \ref{alg:algorithm1} can only find local optimum. This also means that the quality of solution depends on the initialization. In terms of initialization, we can have a GMM by unsupervised Expectation Maximization, which is typically used for computing Fisher vectors. As a matter of fact, it is a common to see that in deep networks layer-wise unsupervised pre-training is widely used for seeking good initialization \cite{HintonFast2006}. To update gradient, we employ a batch setting with a line search (line 10) to find the most effective step size in each iteration. In optimization, the mixture weights $\pi$ and Gaussian variance $\Sigma$ should be ensured to remain positive.
As suggested by \cite{DeepFisherKernel}, this can be achieved by internally reparametering and updating the logarithms of their values, from which
the original parameters can be computed simply by exponentiation. A valid GMM parameterization also requires the mixture weights sum up to 1. Instead of directly enforcing this constraint that would require a projection step after each update, we avoid this by deriving gradients for unnormalized weights $\tilde{\pi}_k$, and then normalize them as $\pi_k = \tilde{\pi}_k/ \sum_j \tilde{\pi}_j$.

\begin{algorithm}[t]
\algorithmicrequire{training samples $X_1,\dots,X_n$, labels $y_1,\dots,y_n$}\\
\algorithmicensure{GMM $G$, parameters $\Theta$ of supervised layers}\\
\textbf{Initialize}: initial GMM, $G=(\log \pi, \mu, \log \Sigma)$ by using unsupervised expectation maximization \cite{FisherVectorIJCV,ImproveFisherKernel}.\\
\Repeat{stopping criterion fulfilled}{
compute Fisher vector with respect to $G$:
$\Phi_i^G = \Phi(x_i; G), i=1,\dots,n$.\\
solve LDA for training set $\{ (\Phi_i^G, y_i), i=1,\dots,n \}$:\\
$\Theta \leftarrow \mathcal{L}(\Theta, G) = \arg\max_{\Theta} \frac{1}{m} \sum_{i=1}^m v_i$ with $\{v_1,\dots,v_m\}=\{v_i | v_i < \min (v_1,\dots,v_{C-1}) + \epsilon \}$.\\
compute gradients with respect to the GMM parameters:\\
$\delta_{\log \pi}=\bigtriangledown_{\log \pi} \mathcal{L}(\Theta, G), \delta_{\mu}=\bigtriangledown_{\mu} \mathcal{L}(\Theta, G), \delta_{\log \Sigma}=\bigtriangledown_{\log \Sigma} \mathcal{L}(\Theta, G).$\\
find the best step size $\eta*$ by line search:\\
$\eta*=\arg\min_{\eta} \mathcal{L}(\Theta, G)$ with $G_{\eta} = (\log \pi -\eta \delta_{\log \pi}, \mu-\eta \delta_{\mu}, \log \Sigma - \eta \delta_{\log \Sigma})$.\\
update GMM parameters: $G \leftarrow G_{\eta*}$.
}
\textbf{Return} GMM parameters $G$ and $\Theta$.
\caption{Deep Fisher learning with linear discriminant analysis.}\label{alg:algorithm1}
\end{algorithm}

\subsection{Implementation details}

\paragraph{Local feature extration} Our architecture is independent of a particular choice of local descriptors. Following \cite{FVsMeetNNs,FisherVectorIJCV}, we combine SIFT with color histogram extracted from the LAB colorspace. Person images are first rescaled to a resolution of $48\times 128$ pixels. SIFT and color features are extracted over a set of 14 dense overlapping $32\times 32$-pixels regions with a step stride of 16 pixels in both directions. For each patch we extract two types of local descriptors: SIFT features and color histograms. For SIFT patterns, we divide each patch into $4\times 4$ cells and set the number of orientation bins to 8, and obtain $4\times 4 \times 8=128$ dimensional SIFT features. SIFT features are extracted in L, A, B color channels and thus produce a $128\times 3$ feature vector for each patch. For color patterns, we extract color features using 32-bin histograms at 3 different scales (0.5, 0.75 and 1) over L, A, and B channels. Finally, the dimensionality of SIFT features and color histograms is reduced with PCA respectively to 77-dim and 45-dim. Then, we concatenate these descriptors with $(x, y)$ coordinates as well as the scale of the paths, thus, yielding 80-dim and 48-dim descriptors.  In our experiments, we use the publicly available SIFT/LAB implementation from \cite{eSDC}. In Fisher vector encoding, we use a GMM with $K=256$ Gaussians. The per-patch Fisher vectors are aggregated with sum-pooling, square-rooted and $\ell_2$-normalized. One Fisher vector is computed on the SIFT and another one on the color descriptors. The two Fisher vectors are concatenated into a $256K$-dim representation.

\section{Comparison with CNNs \cite{AlexNet} and deep Fisher networks \cite{DeepFisherKernel}}\label{sec:relationship}

\paragraph{Comparison with CNNs \cite{AlexNet}} The architecture of deep Fisher kernel learning has substantial number of conceptual similarities with that of CNNs. Like a CNN, the Fisher vector extraction can also be interpreted as a series of layers that alternate linear and non-linear operations, and can be paralleled with the convolutional layers of the CNN. For example, when training a convolutional network on natural images, the first stages always learn the essentially the same functionality, that is, computing local image gradient orientations, and pool them over small spatial regions. This is, however, the same procedure how a SIFT descriptor is computed.
It has been shown in \cite{FVsMeetNNs} that Fisher vectors are competitive with representations learned by the convolutional layers of the CNN, and the accuracy comes close  to that of the standard CNN model, \ie,  AlexNet \cite{AlexNet}.
The main difference between our architecture between CNNs lies in the first layers of feature extractions. In CNNs, basic components (convolution, pooling, and activation functions) operate on local input regions, and locations in higher layers correspond to the locations in the image they are path-connected to (a.k.a receptive field), whilst they are very demanding in computational resources and the amount of training data necessary to achieve good generalization. In our architecture, as alternative, Fisher vector representation of an image effectively encodes each local feature (\eg, SIFT) into a high-dimensional representation, and then aggregates these encodings into a single vector by global sum-pooling over the over image, and followed by normalization. This serves very similar purpose to CNNs, but trains the network at a smaller computational learning cost.

\paragraph{Comparison with deep Fisher kernel learning \cite{DeepFisherKernel}}

With respect to the deep Fisher kernel end-to-end learning, our architecture exhibits two major differences. The first difference can be seen from the incorporation of stacked supervised fully connected layers in our model, which can capture long-range and complex structure on an person image.
The second difference is the training objective function. In \cite{DeepFisherKernel}, they introduce an SVM classifier with Fisher kernel that jointly learns the classifier weights and image representation. However, this classification training can only classify an input image into a number of identities. It cannot be applied to an out-of-domain case where new testing identifies are unseen in training data.  Specifically, this classification supervisory signals pull apart the features of different identities since they have to be classified into different classes whereas the classification signal has relatively week constraint on features extracted from the same identity. This leads to the problem of generalization to new identities in test.  By contrast, we reformulate LDA objective into deep neural network to learn linearly separable representations. In this way, favorable properties of LDA (low (high) intra- (inter-) personal variations  and optimal decision boundaries) can be embedded into neural networks.

\section{Experiments}\label{sec:exp}

\subsection{Experimental setting}

\paragraph{Data sets.} We perform experiments on four benchmarks: VIPeR  \cite{Gray2007Evaluating}, CUHK03 \cite{FPNN}, CUHK01 \cite{GenericMetric} and the Market-1501 data set \cite{Market1501}.

\begin{itemize}
\item The \textbf{VIPeR} data set \cite{Gray2007Evaluating} contains $632$ individuals taken from two cameras with arbitrary viewpoints and varying illumination conditions. The 632 person's images are randomly divided into two equal halves, one for training and the other for testing.

\item The \textbf{CUHK03} data set \cite{FPNN} includes 13,164 images of 1360 pedestrians. The whole dataset is captured with six surveillance camera. Each identity is observed by two disjoint camera views, yielding an average 4.8 images in each view. This dataset provides both manually labeled pedestrian bounding boxes and bounding boxes automatically obtained by running a pedestrian detector \cite{DetectionPAMI}. In our experiment, we report results on labeled data set.

\item The \textbf{CUHK01} data set \cite{GenericMetric}  has 971 identities with 2 images per person in each view. We report results on the setting where 100 identities are used for testing, and the remaining 871 identities used for training, in accordance with FPNN \cite{FPNN}.

\item The \textbf{Market-1501} data set \cite{Market1501} contains 32,643 fully annoated boxes of 1501 pedestrians, making it the largest person re-id dataset to date. Each identity is captured by at most six cameras and boxes of person are obtained by running a state-of-the-art detector, the Deformable Part Model (DPM) \cite{MarketDetector}.  The dataset is randomly divided into training and testing sets, containing 750 and 751  identities, respectively.
\end{itemize}

\paragraph{Evaluation protocol} We adopt the widely used single-shot modality in our experiment to allow extensive comparison. Each probe image is matched against the gallery set, and the rank of the true match is obtained. The rank-$k$ recognition rate is the expectation of the matches at rank $k$, and the cumulative values of the recognition rate at all ranks are recorded as the one-trial Cumulative Matching Characteristice (CMC) results \cite{paul2015ensemble}. This evaluation is performed ten times, and the average CMC results are reported.

\paragraph{Competitors} We compare our model with the following state-of-the-art approaches: SDALF \cite{Farenzena2010Person}, ELF \cite{Gray2008Viewpoint}, LMNN \cite{Hirzer2012Person}, ITML \cite{Davis2007Information}, LDM \cite{Guillaumin2009Isthatyou}, eSDC \cite{eSDC}, Generic Metric \cite{GenericMetric}, Mid-Level Filter (MLF) \cite{MidLevelFilter}, eBiCov \cite{eBiCov}, SalMatch \cite{Zhao2013SalMatch},  PCCA \cite{PCCA}, LADF \cite{LADF}, kLFDA \cite{Xiong2014Person}, RDC \cite{Zheng2013PAMI}, RankSVM \cite{RankSVM}, Metric Ensembles (Ensembles) \cite{paul2015ensemble}, KISSME \cite{Kostinger2012Large}, JointRe-id \cite{JointRe-id}, FPNN \cite{FPNN}, DeepRanking \cite{DeepRanking}, NullReid \cite{NullSpace-Reid} and XQDA \cite{LOMOMetric}. For a fair comparison, the method of Metric Ensembles takes the structured ensembling of two types of features \footnote{In each patch, SIFT features and color histograms are $\ell_2$ normalized to form a discriminative descriptor vector with length $128\times 3$, and $32\times 3 \times 3$, respectively, yielding 5376-dim SIFT and 4032-dim color feature per image. PCA is applied on the two featuers to reduce the dimesionality to be 100-dim, respectively. }, SIFT and LAB patterns, where kLFDA is used as the base metric in its top-$k$ ranking optimization. %

\paragraph{Training} For each image, we form a 6,5536-dimensional Fisher vector based on a 256-component GMM that was obtained prior to learning by expectation maximization. The network has 3 hidden supervised layers with 4096, 1024, and 1024 hidden units and a drop-out rate of 0.2. As we are estimating the distribution of data, we apply batch normalization \cite{BatchNorm} after the last supervised layer. Batch normalization can help to increase the convergence speed and has a positive effect on LDA based optimization. In the training stage, a batch size of 128 is used for all data sets.
The network is trained using SGD with Nesterov momentum. The initial learning rate is set to 0.05 and the momentum is fixed at 0.9. The learning rate is then halved every 50 epochs for CUHK03, Market-1501 data sets, and every 20 epoches for VIPeR and CUHK01 data sets. For further regularization, we add weight decay with a weighting of $10^{-4}$ on all parameters of the model. The between-class covariance matrix regularization weight $\lambda$ (in Eq.\eqref{eq:eigen_prob}) is set to $10^{-3}$ and the $\epsilon$-offset  (in Eq.\eqref{eq:modified_obj}) for LDA optimization is 1.

\subsection{Experimental results}

\begin{table}[t]
  \centering
  \caption{Rank-1 accuracy and number of parameters in training for various deep models on the CUHK03 data set. }\label{tab:compare_model}
  {\begin{tabular}{c|c|c}
  \hline
    Method  & $r=1$  &   \# of parameters \\
  \hline\hline
  DeepFV+CEL & 59.52 & 5,255,268 \\
  VGG+CEL & 58.83 & 22,605,924\\
  VGG+LDA & 63.67 & 22,605,824 \\
  DeepFV+LDA & 63.23 & 5,255,168\\
  \hline
  \end{tabular}
  }\end{table}

In this section, we provide detailed analysis of the proposed architecture on CUHK03 data set. To have a fair comparison, we train a modified model by replacing LDA loss with a cross-entropy loss, denoted as DeepFV+CEL. To validate the parallel performance of Fisher vectors to CNNs, we follow the VGG model \cite{VGG} with sequences of 3$\times$3 convolution, and fine tune its supervised layers to CUHK03 training samples. Two modified VGG models are optimized on LDA loss, and cross-entropy loss, and thus, referred to VGG+LDA, and VGG+CEL, respectively. In training stage of VGG variant models, we perform data pre-processing and augmentation, as suggested in \cite{PersonNet,FPNN}.

Table \ref{tab:compare_model} summarizes the rank-1 recognition rates and number of training parameters in deep models on the CUHK03 data set. It can be seen that high-level connections like convolution and fully connected layers result in a large number of parameters to be learned (VGG+CEL and VGG+LDA). By contrast, hybrid systems with Fisher vectors (DeepFV+CEL, and DeepFV+LDA) can reduce the number of parameters substantially, and thus alleviate the over-fitting to some degree. The performance of DeepFV+LDA is very close to VGG+LDA but with much less parameters in training. Also, deep models with LDA objective outperform those with cross-entropy loss.

\subsection{Comparison with state-of-the-art approaches}

\begin{table}[t]
  \centering
  \caption{Rank-1, -5, -10, -20 recognition rate of various methods on the VIPeR data set.  The method of Ensembles* takes two types of features as input: SIFT and LAB patterns.}  \label{tab:cmc_viper}
  {
  \begin{tabular}{c|c|c|c|c}
\hline
    Method  & $r=1$  &  $r=5$ & $r=10$  & $r=20$ \\
  \hline\hline
   Ensembles* \cite{paul2015ensemble}  & 35.80 & 67.82  & 82.67 & 88.74\\
   JointRe-id \cite{JointRe-id} & 34.80 & 63.32  & 74.79 & 82.45  \\
   LADF \cite{LADF}  & 29.34 & 61.04 & 75.98 & 88.10\\
   SDALF \cite{Farenzena2010Person} & 19.87 & 38.89 & 49.37 & 65.73\\
   eSDC \cite{eSDC} & 26.31 & 46.61 & 58.86 & 72.77\\
   KISSME \cite{Kostinger2012Large} & 19.60 & 48.00 & 62.20 & 77.00\\
   kLFDA \cite{Xiong2014Person} & 32.33 & 65.78 & 79.72 & 90.95\\
  eBiCov \cite{eBiCov} & 20.66 & 42.00 & 56.18 & 68.00 \\
  ELF \cite{Gray2008Viewpoint} & 12.00 & 41.50 & 59.50 & 74.50\\
  PCCA \cite{PCCA} & 19.27 & 48.89 & 64.91 & 80.28\\
  RDC \cite{Zheng2013PAMI} & 15.66 & 38.42 & 53.86 & 70.09\\
  RankSVM \cite{RankSVM} & 14.00 & 37.00 & 51.00 & 67.00\\
  DeepRanking \cite{DeepRanking} & 38.37 & 69.22 & 81.33 & 90.43\\
 NullReid \cite{NullSpace-Reid} & 42.28 & 71.46 & $\mathbf{82.94}$ & $\mathbf{92.06}$ \\
\hline
   Hybrid   &  $\mathbf{44.11}$  & $\mathbf{72.59}$ & 81.66 &  91.47\\
  \hline
  \end{tabular}
  }
\end{table}

\begin{table}[t]
  \centering
  \caption{Rank-1, -5, -10, -20 recognition rate of various methods on the CUHK03 data set.}  \label{tab:cmc_cuhk03}
  {
  \begin{tabular}{c|c|c|c|c}
\hline
    Method  & $r=1$  &  $r=5$ & $r=10$  & $r=20$ \\
  \hline\hline
   Ensembles* \cite{paul2015ensemble}  & 52.10 &  59.27 & 68.32 & 76.30\\
   JointRe-id \cite{JointRe-id}  & $54.74$ & 86.42 & 91.50 & 97.31 \\
   FPNN \cite{FPNN} & $20.65$ & 51.32 & 68.74 & 83.06\\
   NullReid \cite{NullSpace-Reid}  & $58.90$ & 85.60 & $92.45$ & 96.30 \\
   ITML \cite{Davis2007Information} & $5.53$ & 18.89 &29.96 & 44.20 \\
   LMNN \cite{Hirzer2012Person} & $7.29$ & 21.00 & 32.06& 48.94 \\
   LDM \cite{Guillaumin2009Isthatyou} & $13.51$ & 40.73 &52.13 & 70.81 \\
   SDALF \cite{Farenzena2010Person} & $5.60$ & 23.45 &36.09 & 51.96 \\
   eSDC \cite{eSDC} & $8.76$ & 24.07 &38.28 & 53.44 \\
   KISSME \cite{Kostinger2012Large} & $14.17$ & 48.54 &52.57 & 70.03 \\
   kLFDA \cite{Xiong2014Person} & 48.20 & 59.34 & 66.38 & 76.59\\
   XQDA \cite{LOMOMetric} & 52.20 & 82.23 & 92.14 & 96.25\\
\hline
   Hybrid   &  $\mathbf{63.23}$ &  $\mathbf{89.95}$ & $\mathbf{92.73}$ &  $\mathbf{97.55}$ \\
  \hline
  \end{tabular}
  }
\end{table}

\begin{table}[t]
  \centering
  \caption{Rank-1, -5, -10, -20 recognition rate of various methods on the CUHK01 data set.}  \label{tab:cmc_cuhk01}
  {
  \begin{tabular}{c|c|c|c|c}
  \hline
\hline
    Method  & $r=1$  & $ r=5 $& $r=10 $ & $r=20$ \\
  \hline\hline
   Ensembles* \cite{paul2015ensemble}  & 46.94 &  71.22 & 75.15 & 88.52\\
   JointRe-id \cite{JointRe-id}  & $65.00$ & $88.70$ & $\mathbf{93.12}$ & $\mathbf{97.20}$ \\
   SDALF \cite{Farenzena2010Person} & 9.90 & 41.21 & 56.00 & 66.37 \\
   MLF \cite{MidLevelFilter} & 34.30 & 55.06 & 64.96 & 74.94\\
   FPNN \cite{FPNN} & 27.87 & 58.20 & 73.46 & 86.31 \\
   LMNN \cite{Hirzer2012Person} & 21.17 & 49.67 & 62.47 & 78.62 \\
   ITML \cite{Davis2007Information} & 17.10 & 42.31 & 55.07 & 71.65 \\
   eSDC \cite{eSDC} & 22.84 & 43.89 & 57.67 & 69.84 \\
   KISSME \cite{Kostinger2012Large}  & 29.40 & 57.67 & 62.43 & 76.07 \\
   kLFDA \cite{Xiong2014Person} & 42.76 & 69.01 & 79.63 & 89.18\\
  Generic Metric \cite{GenericMetric} & 20.00 & 43.58 & 56.04 & 69.27\\
   SalMatch \cite{Zhao2013SalMatch} & 28.45 & 45.85 & 55.67 & 67.95\\
\hline
Hybrid & $\mathbf{67.12}$ & $\mathbf{89.45}$ & 91.68 & $96.54$ \\
  \hline
  \end{tabular}
  }
\end{table}

\begin{table}[t]
  \centering
  \caption{Rank-1 and  mAP of various methods on the Market-1501 data set.} \label{tab:cmc_market}
  {
  \begin{tabular}{c|c|c}
\hline
    Method  & $ r=1$  &  mAP \\
  \hline\hline
   SDALF \cite{Farenzena2010Person} & 20.53 & 8.20\\
   eSDC \cite{eSDC} & 33.54 & 13.54\\
   KISSME \cite{Kostinger2012Large} & 39.35 & 19.12\\
   kLFDA \cite{Xiong2014Person} & 44.37 &  23.14 \\
   XQDA \cite{LOMOMetric} & 43.79 & 22.22\\
   Zheng \etal \cite{Market1501} & 34.40 &  14.09\\
\hline
Hybrid & $\mathbf{48.15}$ & $\mathbf{29.94}$ \\
  \hline
  \end{tabular}
  }
\end{table}

\begin{figure}[h!]
    \centering
     \includegraphics[width=0.458\textwidth]{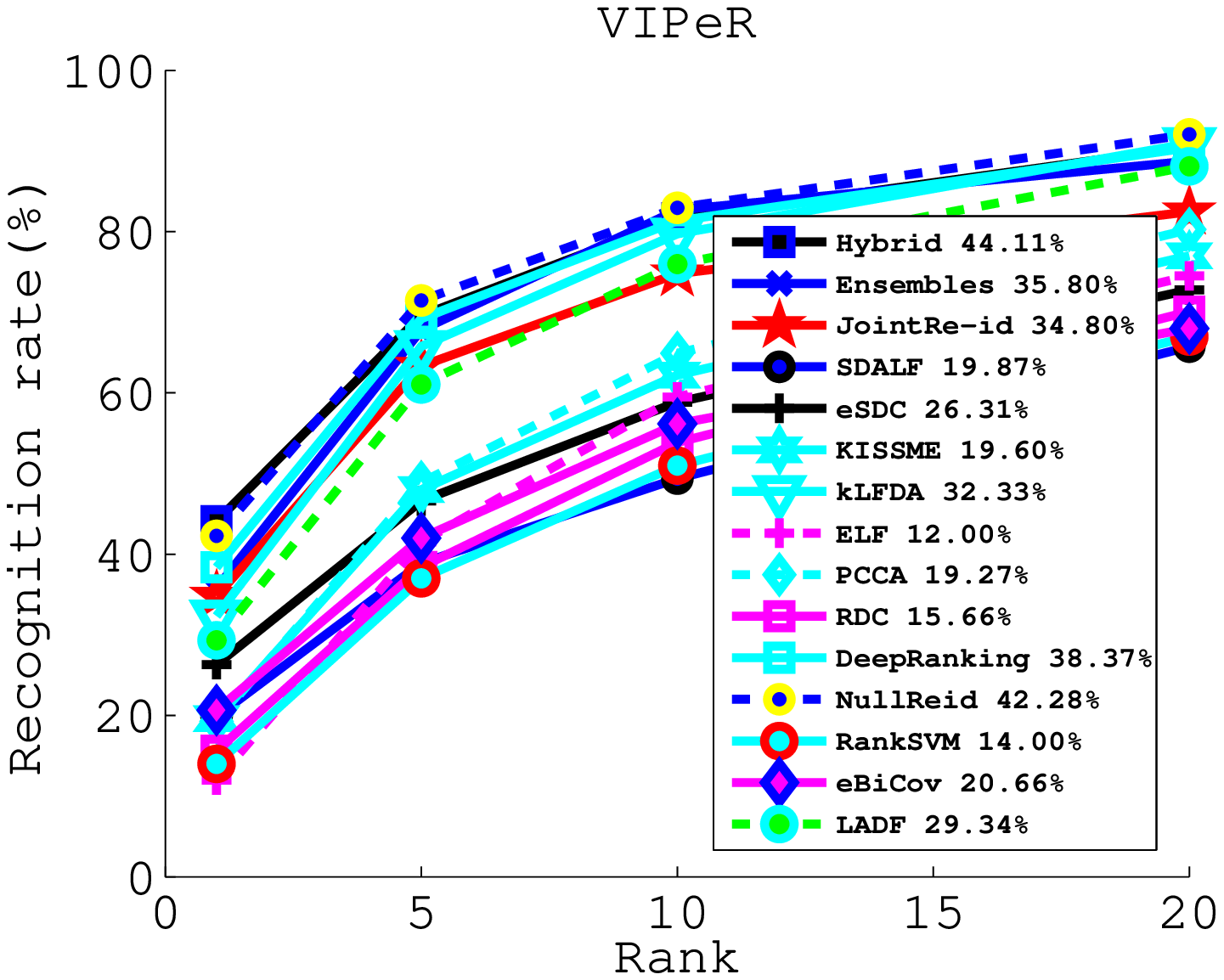}
	   \includegraphics[width=0.458\textwidth]{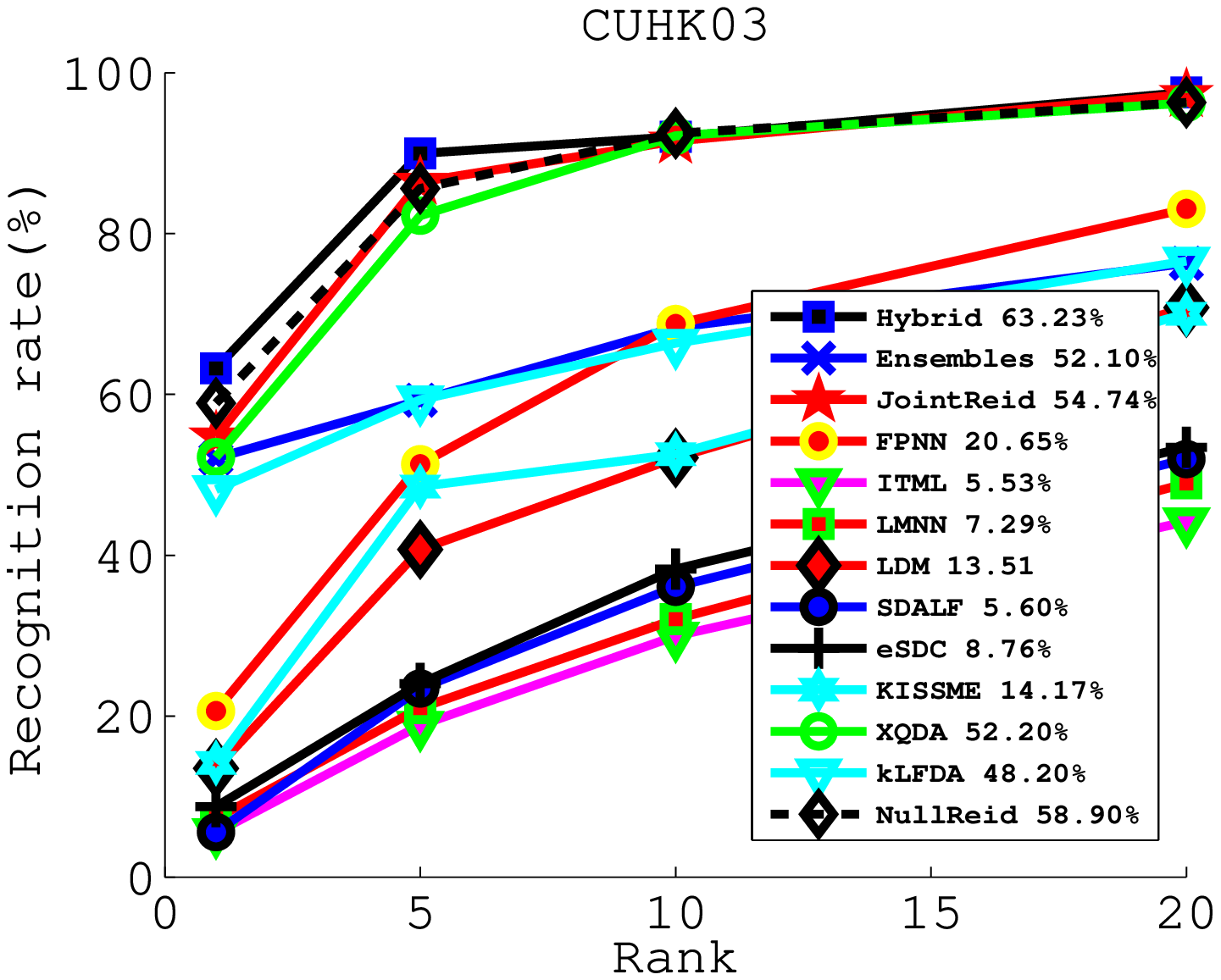}
     \includegraphics[width=0.458\textwidth]{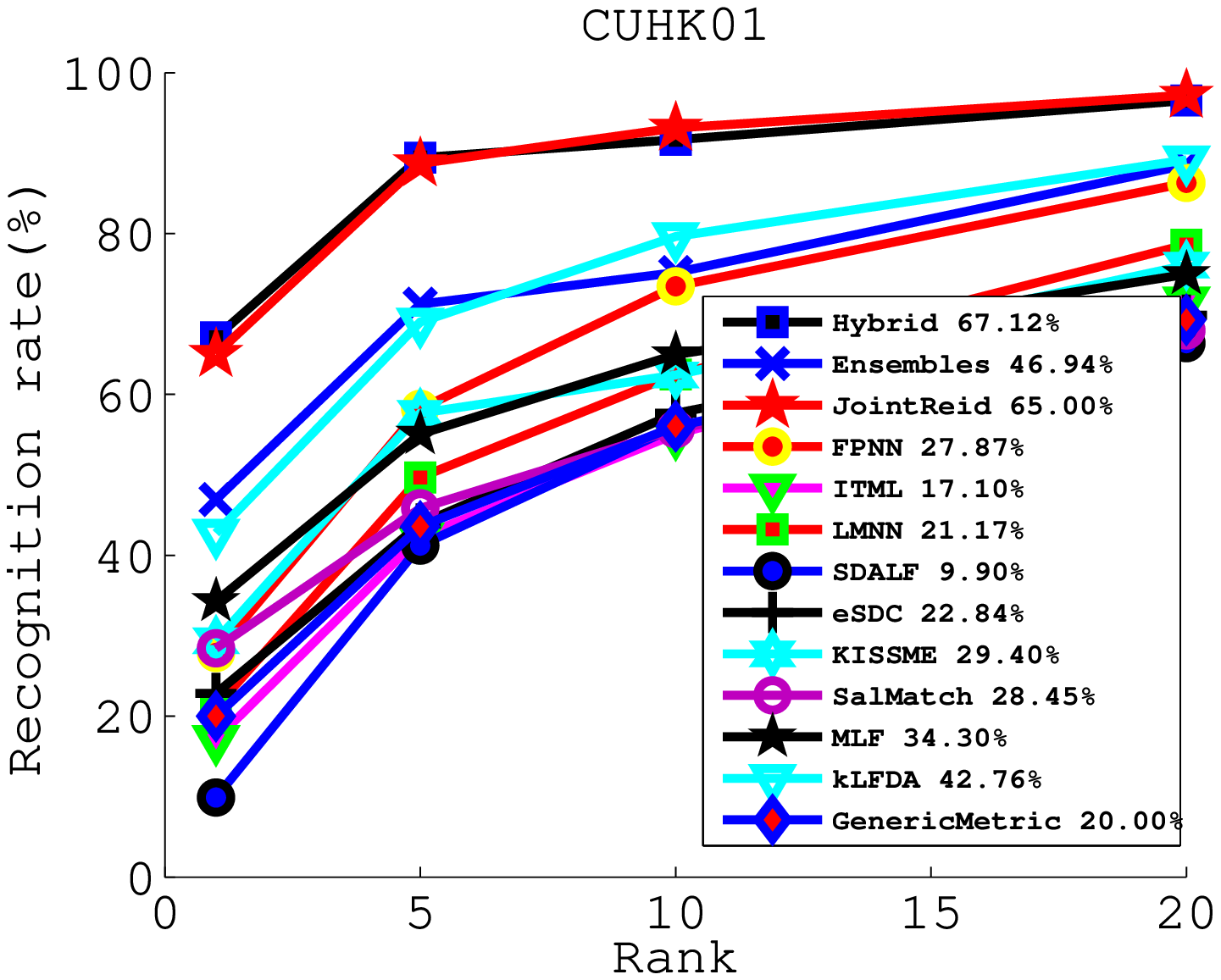}
    \caption{Performance comparison with state-of-the-art approaches using CMC curves on VIPeR, CUHK03 and CUHK01 data sets.}\label{fig:cmc}
\end{figure}

\subsubsection{Evaluation on the VIPeR data set}
Fig. \ref{fig:cmc} (a) shows the CMC curves up to rank-20 recognition rate comparing our method with state-of-the-art approaches. It is obvious that our mothods delivers the best result in rank-1 recognition. To clearly present the quantized comparison results, we summarize the comparison results on seveval top ranks in Table \ref{tab:cmc_viper}. Sepcifically, our method achieves $44.11\%$ rank-1 matching rate, outperforming the previous best results $42.28\%$ of NullReid \cite{NullSpace-Reid}. Overall, our method performs best over rank-1, and 5, whereas NullReid is the best at rank-15 and 20. We suspect that training samples in VIPeR are much less to learn parameters of moderate size in our deep hybrid model.

\subsubsection{Evaluation on the CUHK03 data set}

The CUHK03 data set is larger in scale than VIPeR. We compare our model with state-of-the-art approaches including deep learning methods (FPNN \cite{FPNN}, JointRe-id \cite{JointRe-id}) and metric learning algorithms (Ensembles \cite{paul2015ensemble}, ITML \cite{Davis2007Information}, LMNN \cite{Hirzer2012Person}, LDM \cite{Guillaumin2009Isthatyou}, KISSME \cite{Kostinger2012Large}, kLFDA \cite{Xiong2014Person}, XQDA \cite{LOMOMetric}). As shown in Fig. \ref{fig:cmc} (b) and Table \ref{tab:cmc_cuhk03}, our methods outperforms all competitors against all ranks. In particular, the hybrid architecture achieves a rank-1 rate of $63.23\%$, outperforming the previous best result reported by JointRe-id \cite{JointRe-id} by a noticeable margin. This marginal improvement can be attributed to the availability of more training data that are fed to the deep network to learn a data-dependent solution for this specific senario. Also, CUHK03 provides more shots for each person, which is beneficial to learn feature representations with discriminative distribution parameters (within and between class scatter).

\subsubsection{Evaluation on the CUHK01 data set}

In this experiment, 100 identifies are used for testing, with the remaining 871 identities used for training. As suggested by FPNN \cite{FPNN} and JointRe-id \cite{JointRe-id}, this setting is suited for deep learning because it uses 90\% of the data for training. Fig. \ref{fig:cmc} (c) compares the performance of our network with previous methods. Our method outperforms the state-of-the-art in this setting with 67.12\% in rank-1 matching rate (vs. 65\% by the next best method).

\subsubsection{Evaluation on the Market-1501 data set}

It is noteworthy that VIPeR, CUHK03, and CUHK01 data sets use regular hand-cropped boxes in the sense that pedestrians are rigidly enclosed under fixed-size bounding boxes, while Market-1501 employs the detector of Deformable Part Model (DPM) \cite{DetectionPAMI} such that images undergo extensive pose variation and misalignment. Thus, we conduct experiments on this data set to verify the effectiveness of the proposed method in more realistic situation. Table \ref{tab:cmc_market} presents the results, showing that our mothod outperforms all competitors in terms of rank-1 matching rate and mean average precision (mAP).

\subsection{Discussions}

\subsubsection{Compare with kernel methods}

\begin{figure*}[hbt]
    \centering
        \includegraphics[width=2.2in,height=1.8in]{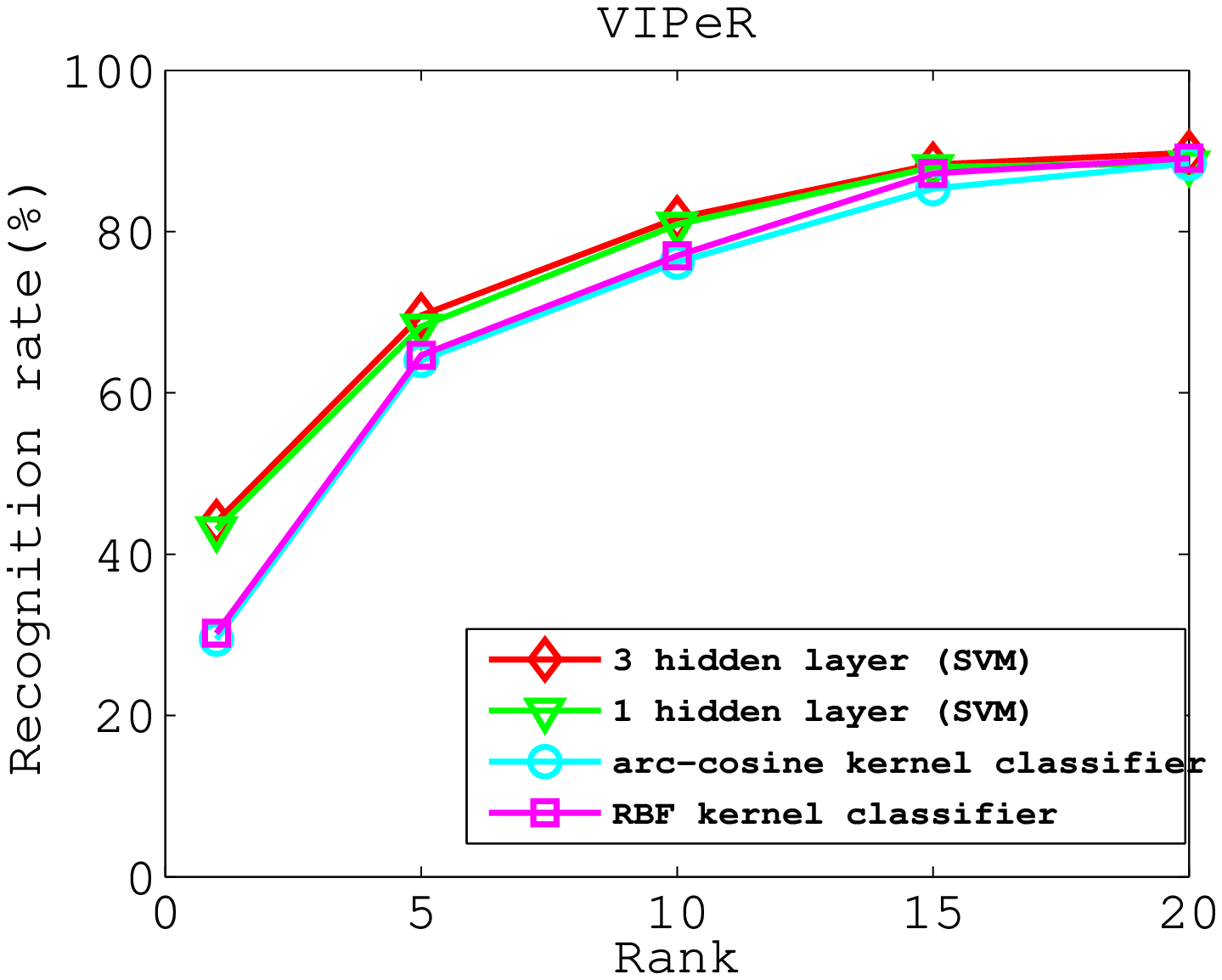}
	\includegraphics[width=2.2in,height=1.8in]{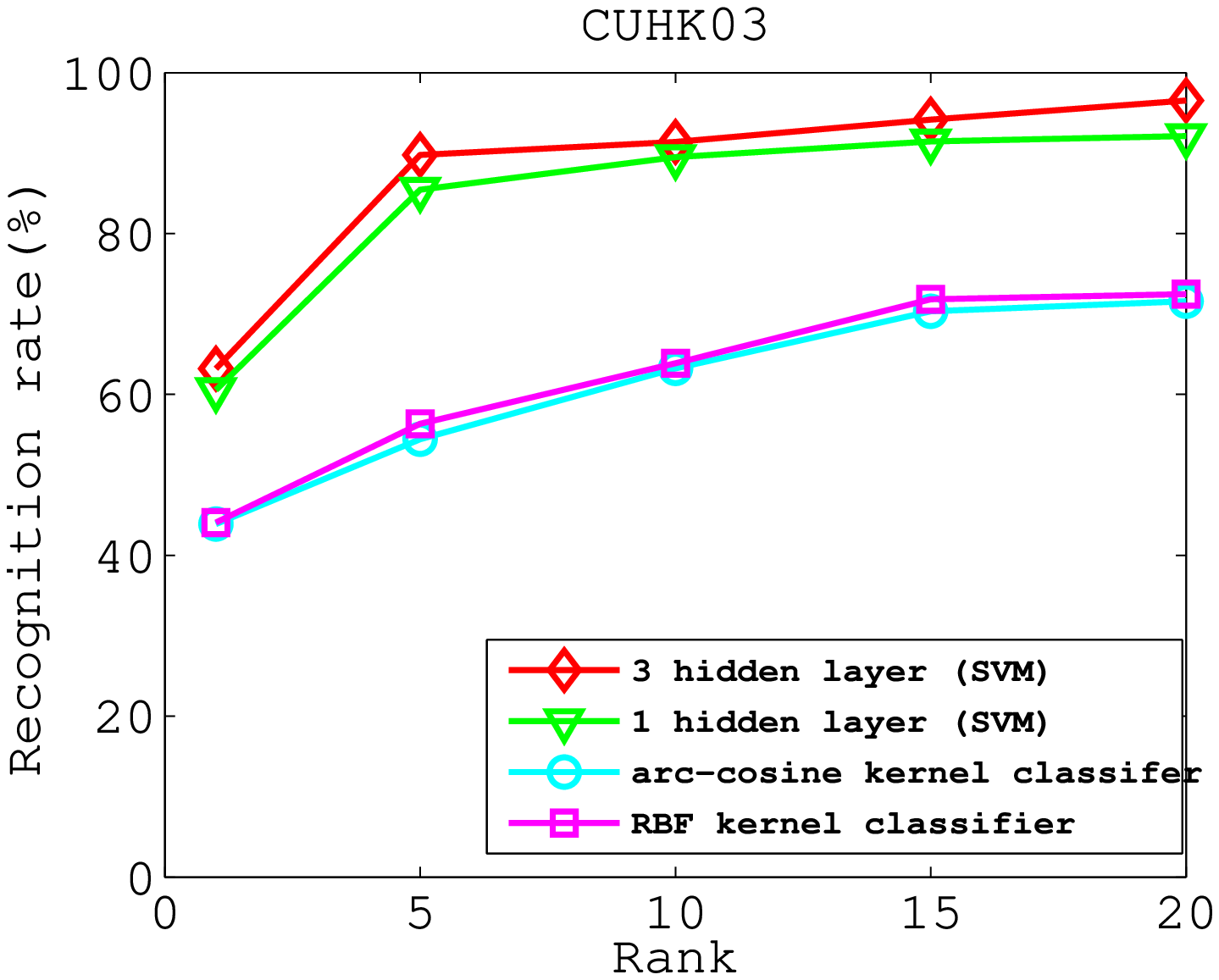}
        \includegraphics[width=2.2in,height=1.8in]{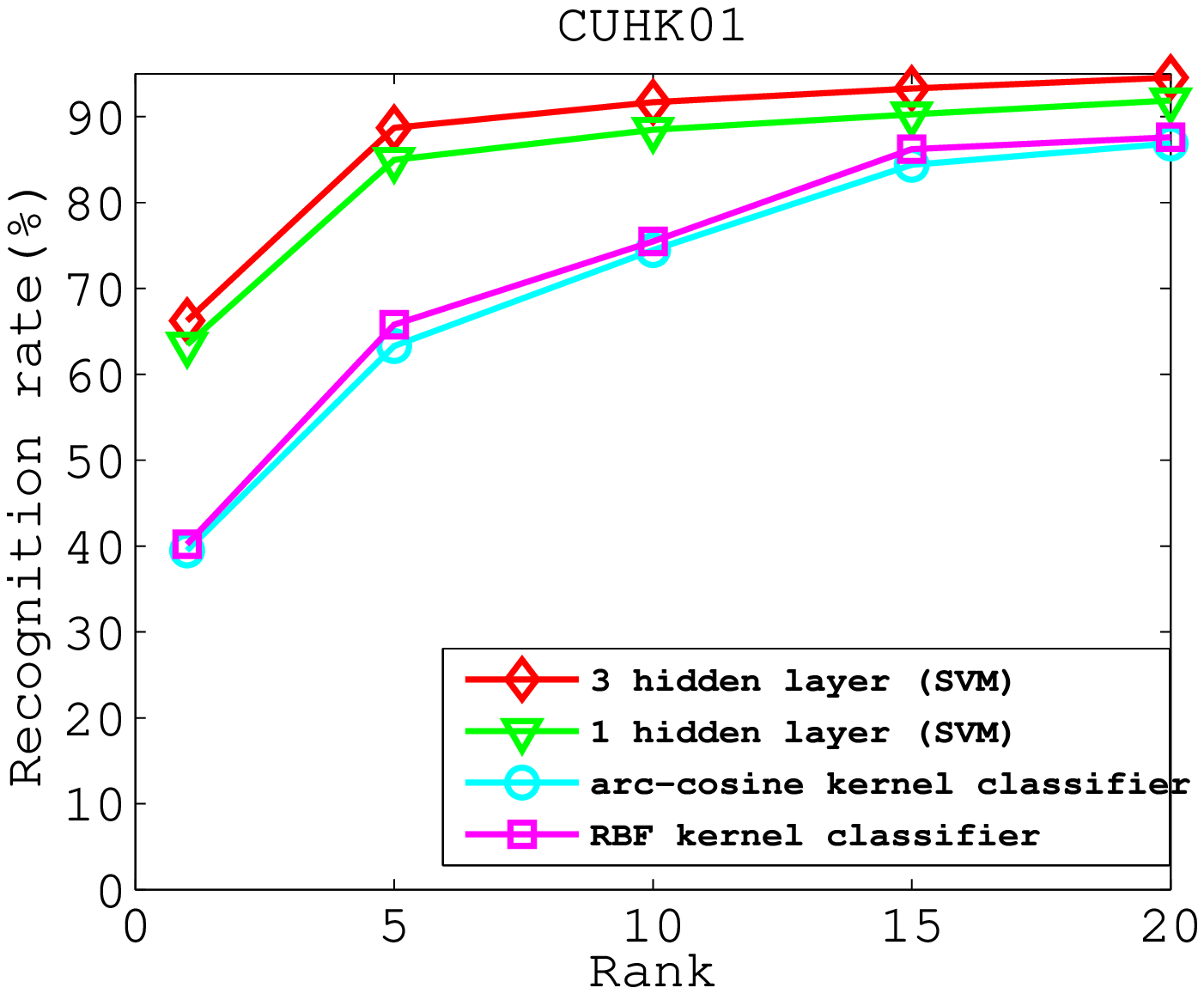}\\
    \caption{Compare with kernel methods using CMC curves on VIPeR, CUHK03 and CUHK01 data sets.}\label{fig:compare_kernel}
\end{figure*}

In the standard Fisher vector classification pipeline, supervised learning relies on kernel classifiers \cite{FisherVectorIJCV} where a kernel/linear classifier can be learned in the non-linear embeddding space. We remark that our architecture with a single hidden layer has a close link to kernel classifiers in which the first layer can be interpreted as a non-linear mapping $\Psi$ followed by a linear classifier learned in the embedding space.  For instance, $\Psi$ is the feature map of arc-cosine kernel of order one if $\Psi$ instantiates random Gaussian projections followed by an reLU non-linearity \cite{KernelDeep}. This shows a close connection between our model (with a single hidden layer and an reLU non-linearity), and arc-cosine kernel classifiers. Therefore, we compare two alternatives with arc-cosine kernel and RBF kernel. Following \cite{FVsMeetNNs}, we train the non-linear kernel classifiers in the primal by leveraging explicit feature maps. For arc-cosine kernel, we use the feature map involving a random Normal projection followed by an reLU non-linearity. For the RBF kernel, we employ the explicit feature map involving a random Normal projection, a random offset and a cosine non-linearity. For our model, we extract feature outputs of the first and the third hidden layer for which a SVM classifier is trained, respectively. Results are shown in Fig. \ref{fig:compare_kernel} as a function of rank-$r$ recognition rate. The arc-cosine and RBF kernels perform similarly and worse than our model with a single hidden layer. The SVM classifier trained on the features of three hidden layers consistently achieves superior performance because this latent space is the most linearly separable.

\subsubsection{Eigenvalue structure of LDA representations}

\begin{figure*}[hbt]
    \centering
        \includegraphics[width=2.2in,height=1.1in]{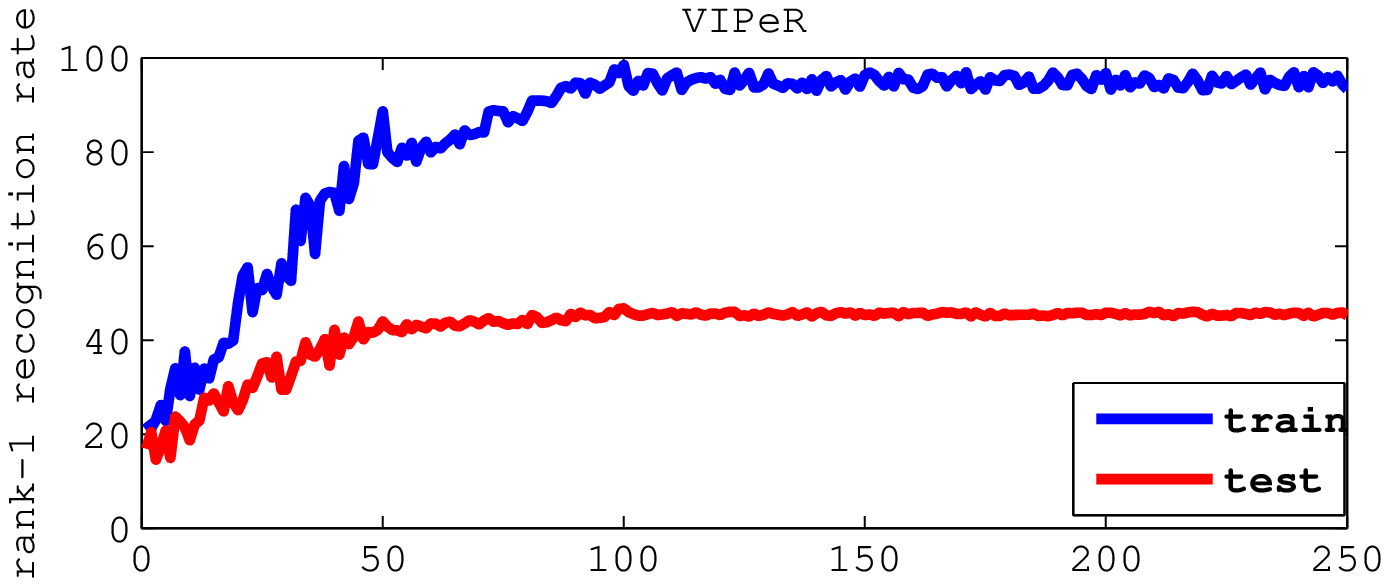}
	\includegraphics[width=2.2in,height=1.1in]{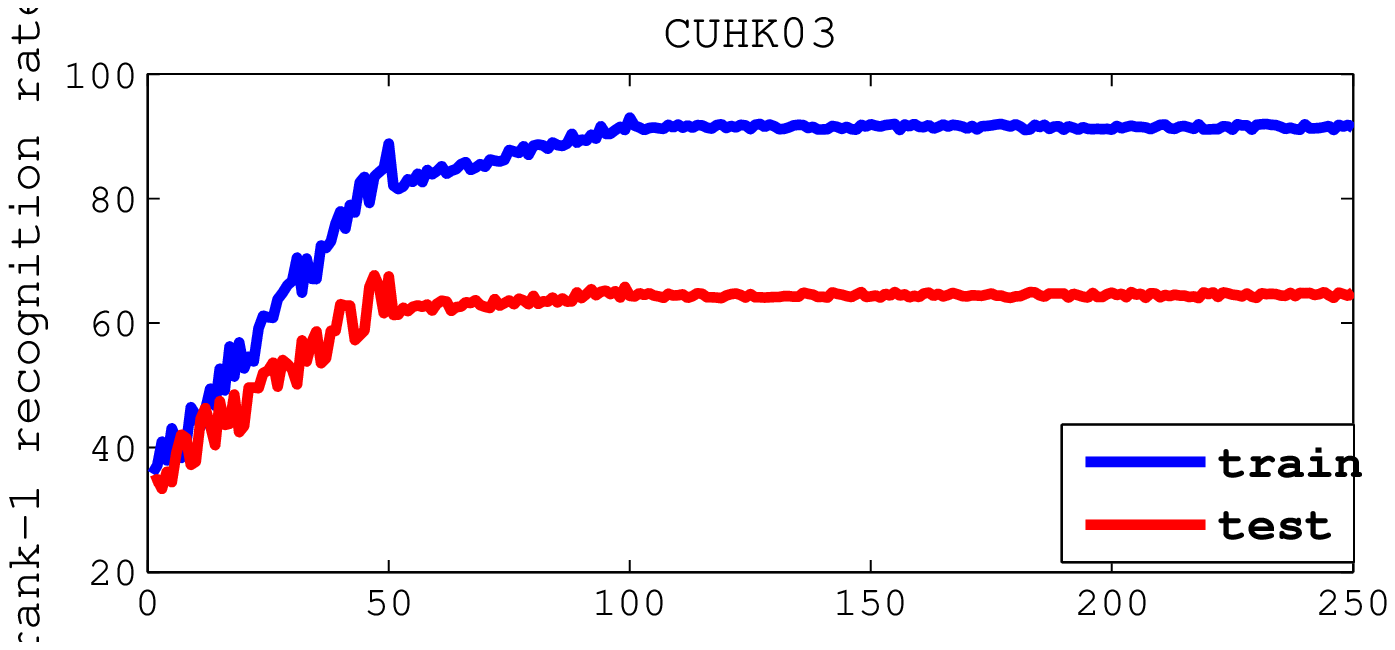}
        \includegraphics[width=2.2in,height=1.1in]{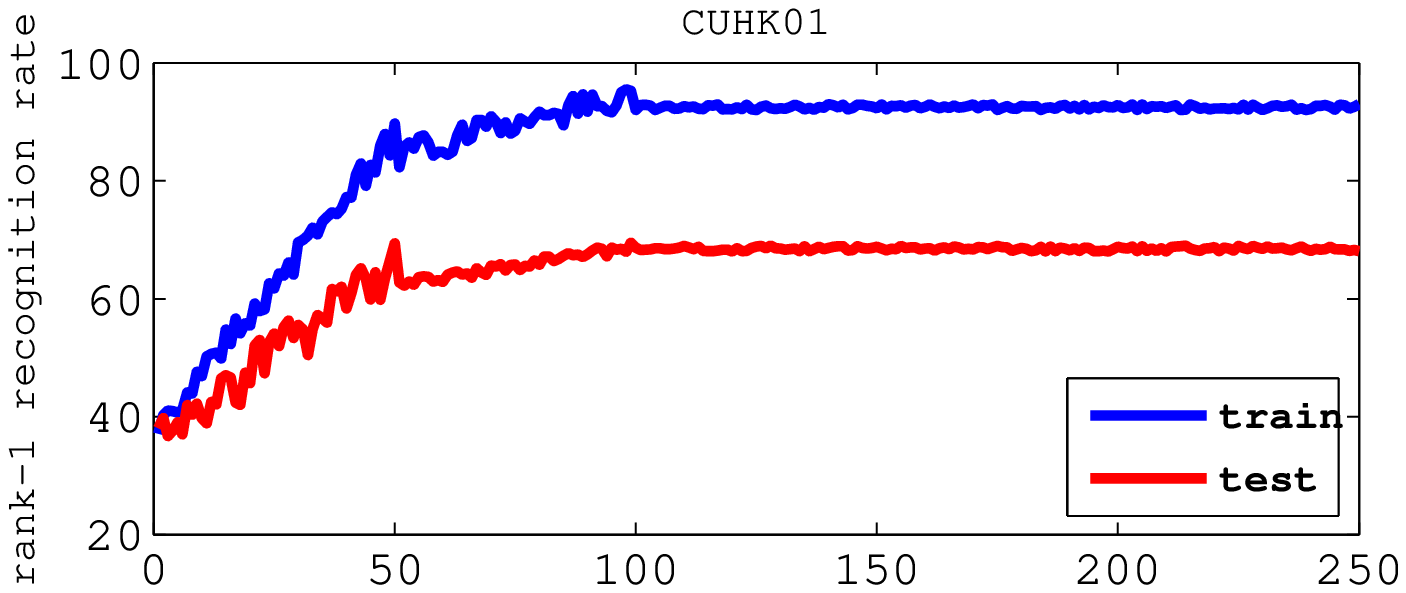}\\
        \includegraphics[width=2.2in,height=1.1in]{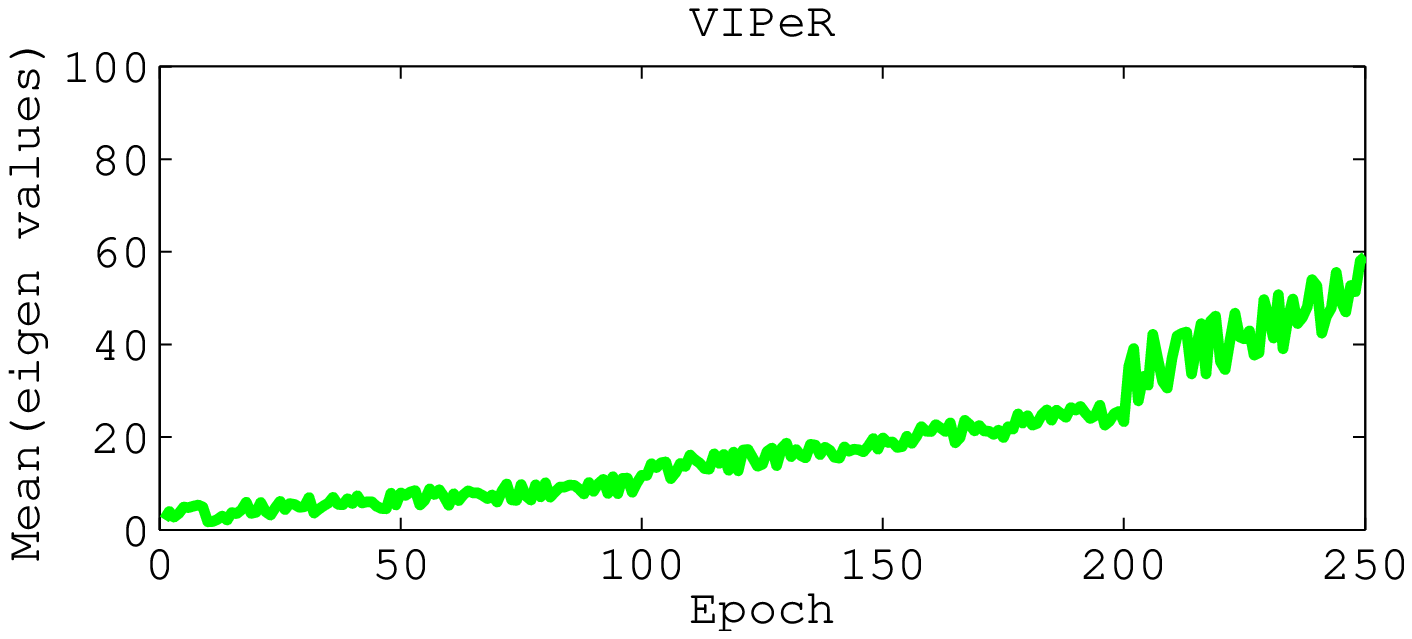}
	\includegraphics[width=2.2in,height=1.1in]{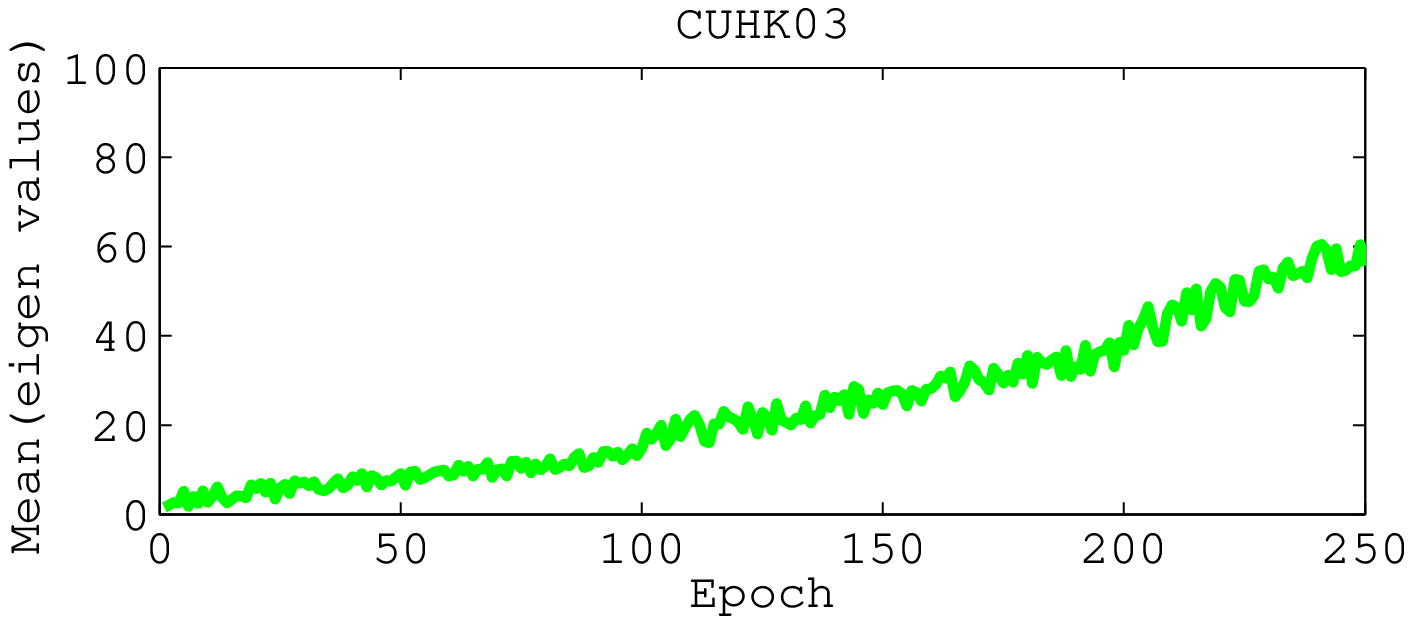}
        \includegraphics[width=2.2in,height=1.1in]{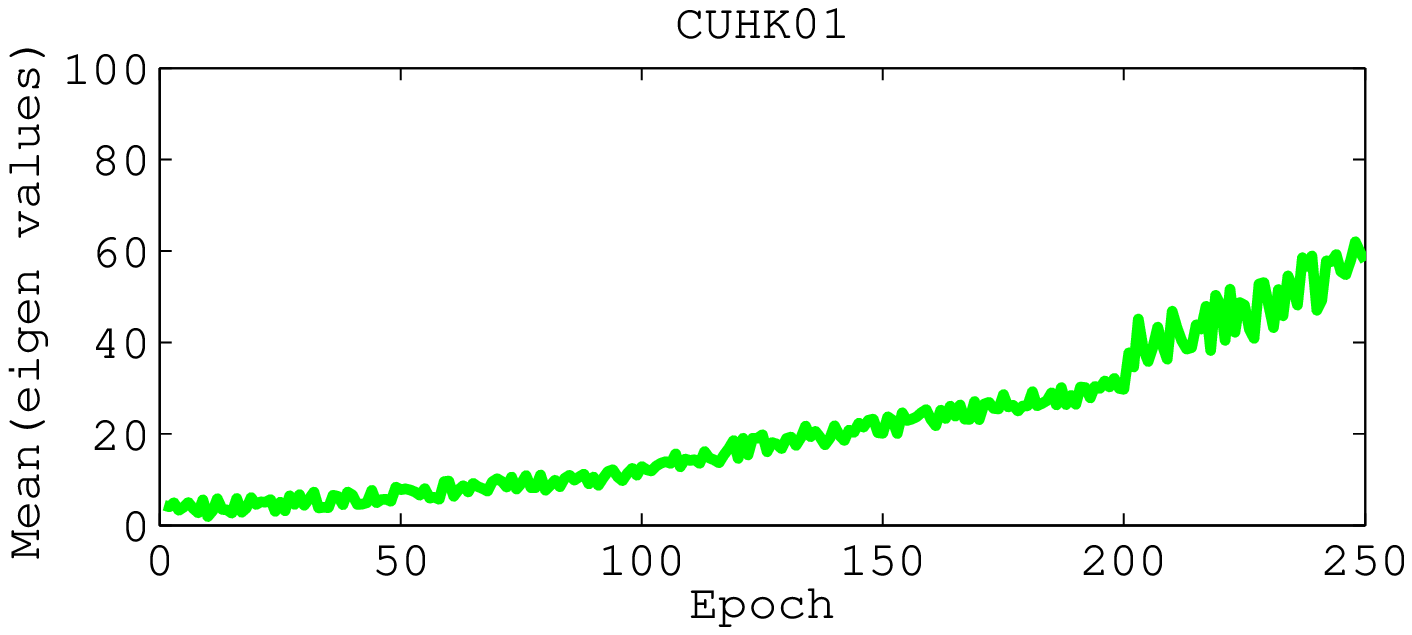}\\
    \caption{Study on eigenvalue structure of LDA representations. Each column corresponding to a benchmark shows the evolution of rank-1 recognition rate along with magnitude of discriminative separation in the latent representations.}\label{fig:eigenvalues}
\end{figure*}

Recall that the eigenvalues derived from the hidden representations can account for the extent to which the discriminative variance in the direction of the corresponding eigenvectors. In this experiment, we investigate the eigenvalue structure of the general LDA eigenvalue problem. Fig. \ref{fig:eigenvalues} show the evolution of training and test rank-1 recognition rate of three data sets along with the mean value of eigenvalues in the training epoch. It can be seen that the discriminative potential of the resulting representation is correlated to the magnitude of eigenvalues on the three data sets. This emphasizes the essence of LDA objective function which allows to embed discriminability into the lower-dimensional eigen-space.

\section{Conclusion}\label{sec:con}

In this paper, we presented a novel deep hybrid architecture to person re-identification. The proposed network composed of Fisher vector and deep neural networks is designed to produce linearly separable hidden representations for pedestrian samples with extensive appearance variations. The network is trained in an end-to-end fashion by optimizing a Linear Discriminant Analysis (LDA) eigen-valued based objective function where LDA based gradients can be back-propagated to update parameters in Fisher vector encoding. We demonstrate the superiority of our method by comparing with state of the arts on four benchmarks in person re-identification.
\section{Appendix}\label{sec:appendix}

This appendix provides explicit expression for the gradients of loss function $\mathcal{L}$ w.r.t. the GMM components. 

Let $k, k'= 1,\dots,K$ be all GMM component indicies, and $d,d'= 1,\dots,D$ be vector component indicies, the derivatives of the Fisher vectors in Eq.\eqref{eq:FV} can be derived as
\begin{equation}\small
\begin{split}
&\frac{\partial \varphi_{d'}^{k'}}{\partial \pi^k}=\frac{\gamma_{k'} \alpha_{d'}^{k'}}{2\pi^k \sqrt{\pi^{k'}}} (\pi^k + \delta_{kk'} -2\gamma_k)\\
&\frac{\partial \varphi_{d'}^{k'}}{\partial \mu^k_d}=\frac{\gamma_{k'} }{\sigma_d^k \sqrt{\pi^{k'}}} (\alpha_{d'}^{k'} \alpha_d^k (\delta_{kk'}-\gamma_k)-\delta^{kk'}_{dd'} )\\
&\frac{\partial \varphi_{d'}^{k'}}{\partial \sigma^k}=\frac{\gamma_{k'} \alpha_{d'}^{k'}}{\sigma_d^k \sqrt{\pi^{k'}}} \left( ( (\alpha_d^k)^2 -1)(\delta_{kk'}-\gamma_k)-   \delta^{kk'}-{dd'}  \right)\\
&\frac{\partial \psi_{d'}^{k'}}{\partial \pi^k}=\frac{\gamma_{k'} ( (\alpha_{d'}^{k'})^2 -1)}{2\pi^k \sqrt{2\pi^{k'}}} (\pi^k + \delta_{kk'} -2\gamma_k)\\
&\frac{\partial \psi_{d'}^{k'}}{\partial \mu^k_d}=\frac{\gamma_{k'} \alpha_d^k}{\sigma_d^k  \sqrt{2\pi^{k'}}} \left( ((\alpha_{d'}^{k'})^2 -1) (\delta_{kk'}-\gamma_k)- 2\delta^{kk'}_{dd'} \right)\\
&\frac{\partial \psi_{d'}^{k'}}{\partial \sigma^k_d}=\frac{\gamma_{k'} }{\sigma_d^k \sqrt{2\pi^{k'}}} \left( ((\alpha_{d'}^{k'})^2 -1) ((\alpha_d^k)^2 -1) (\delta_{kk'}-\gamma_k)- 2\delta^{kk'}_{dd'} (\alpha_d^k)^2\right)\\
\end{split}
\end{equation}
where $\alpha^k :=\frac{x-\mu^k}{\sigma^k}$, and $\delta_{ab}=1$ if $a=b$, and 0 otherwise, and $\delta_{cd}^{ab}=\delta_{ab}\delta_{cd}$.

Stacking the above expressions and averaging them over all descriptors in an image $X$, we obtain the gradient of the unnormalized per-image Fisher vector $\bigtriangledown\Phi(X)$. Then the gradient of the normalized Fisher vector in Eq.\eqref{eq:fisher_norm} can be computed by the chain rule:
\begin{equation}\small
\bigtriangledown \bar{\Phi}_d = \left( \frac{\bigtriangledown \Phi_d}{2\Phi_d} - \frac{\sum_{d'} sign (\Phi_{d'})  \bigtriangledown \Phi_{d'}}{2||\Phi||_{L^1}}\right) \bar{\Phi}_d,
\end{equation}
where the gradients act w.r.t. all parameters, $G=(\pi,\mu,\Sigma)$. Hence, the gradient of the loss term follows by applying the chain rule:
\begin{equation}\label{eq:gradients}\scriptsize
\begin{split}
&\frac{\partial}{\partial \bX}\frac{1}{m}\sum_{i=1}^m v_i =\frac{1}{m}\sum_{i=1}^m \frac{\partial v_i}{\partial \bX}\\
&=\frac{1}{m} \sum_{i=1}^m \be_i^T \left( \frac{\partial \bS_b}{\partial \bX} \cdot \frac{\partial \bX}{\partial \Phi_d} \cdot \bigtriangledown\bar{\Phi}_d - \frac{\partial \bS_w}{\partial \bX} \cdot \frac{\partial \bX}{\partial \Phi_d} \cdot \bigtriangledown\bar{\Phi}_d \right) \be_i.
\end{split}
\end{equation}

The gradients computed in Eq.\eqref{eq:gradients} allow us to apply backpropagation, and train the hybrid network in an end-to-end fashion. Also, a training sample $X$ goes feed-forwadly through the network by Fisher vector encoding $\Phi_G(X)$ (parameterized by $G$), and supervised layers $g_\Theta(\Phi_G(X))$ (parameterized by $\Theta$), such that the hidden representation can be outputed $\bX$ accordingly.
Although computing the gradients with the above expressions is computationally costly, we are lucky to have some strategies to accelerate it. For instance, as each expression contains a term $\gamma_{k'}$, it is suggested to compute the gradient terms only if this value exceeds a threshold \cite{DeepFisherKernel}, \eg $10^{-5}$. Another promising speedup can be achived by subsampling the number of descriptors used from each image to form the gradient. In our experiments, we only use a fraction of 10\% without noticeable loss, however.

\ifCLASSOPTIONcaptionsoff
  \newpage
\fi

\bibliographystyle{IEEEtran}
\bibliography{CSRef}

\end{document}